%% file: main.tex
\RequirePackage{fix-cm}
\documentclass[twocolumn]{svjour3}          
\smartqed  
\usepackage{graphicx}
\usepackage{amsmath}
\usepackage{multirow}
\usepackage{pifont}
\usepackage{graphicx}
\usepackage{amsmath}
\usepackage{amssymb}
\usepackage{array}
\usepackage{multirow, nicefrac}
\usepackage{cite}
\usepackage[caption=false,font=normalsize,labelfont=sf,textfont=sf]{subfig}

\usepackage{color, xcolor}
\usepackage{multirow}
\usepackage{makecell}
\usepackage{multicol}
\usepackage{bbm}
\usepackage{xcolor}
\usepackage{xspace}
\usepackage{float}
\usepackage{color}
\usepackage{colortbl}
\usepackage[misc]{ifsym}
\usepackage{dsfont}
\usepackage[margin=4pt,font=small,labelfont=bf,labelsep=endash,tableposition=top]{caption}
\usepackage{hyperref}

\include{math_commands}

\newcolumntype{I}{!{\vrule width 3pt}}

\newcommand{\textvr}{\textup{TextVR}\xspace}
\newcommand{\starvr}{\textup{StarVR}\xspace}
 
\renewcommand{\texttt}[1]{ $ {{\tt #1} } $}  
\usepackage{dsfont}

\newcolumntype{I}{!{\vrule width 3pt}}
\newlength\savedwidth
\newcommand\whline{\noalign{\global\savedwidth\arrayrulewidth
                           \global\arrayrulewidth 2pt}%
                  \hline
                  \noalign{\global\arrayrulewidth\savedwidth}}
\newlength\savewidth
\newcommand\shline{\noalign{\global\savewidth\arrayrulewidth
                           \global\arrayrulewidth 0.5pt}%
                  \hline
                  \noalign{\global\arrayrulewidth\savewidth}}

\usepackage[margin=4pt,font=small,labelfont=bf,labelsep=endash,tableposition=top]{caption}

\def\eg{\emph{e.g., }}

\def\ie{\emph{i.e., }}

\newcommand{\green}[1]{\textcolor[RGB]{96,177,87}{#1}}

\definecolor{mygray}{gray}{.92}

\definecolor{demphcolor}{RGB}{144,144,144}
\newcommand{\demph}[1]{\textcolor{demphcolor}{#1}}

\definecolor{green}{rgb}{1,0,0}

\newcolumntype{I}{!{\vrule width 3pt}}

\definecolor{mygray}{gray}{.92}

\def\eg{\emph{e.g., }}

\def\ie{\emph{i.e., }}

\usepackage{listings,xcolor} 
\usepackage{tcolorbox}

\begin{document}

\title{A Large Cross-Modal Video Retrieval Dataset with Reading Comprehension
}

\author{Weijia Wu, Yuzhong Zhao, Zhuang Li, Jiahong Li, Hong Zhou, Mike Zheng Shou, and Xiang Bai
}
\authorrunning{W. Wu, Y. Zhao, Z. Li, J. Li, H. Zhou, M. Shou, X. Bai} 

\institute{
Weijia Wu, Hong Zhou({\color{blue}{\Letter}})\at
              Zhejiang University \\
              \email{weijiawu@zju.edu.cn,
              zhouh@mail.bme.zju.edu.cn}
           \and
           Yuzhong Zhao \at University of Chinese Academy of Sciences
               \\    \and
           Zhuang Li, Jiahong Li \at
              Kuaishou Technology\\ 
              \and
           Mike Zheng Shou \at
              National University of Singapore, Singapore\\ 
              \and
           Xiang Bai \at
              Huazhong University of Science and Technology
}

\date{\today}

\maketitle

\begin{abstract}
Most existing cross-modal language-to-video retrieval~(VR) research focuses on single-modal input from video, \ie{} visual representation, while the text is omnipresent in human environments and frequently critical to understand video.
To study how to retrieve video with both modal inputs, \ie{} visual and text semantic representations, we firstly introduce a large-scale and cross-modal \textbf{V}ideo \textbf{R}etrieval dataset with text reading comprehension, \textvr, which contains $42.2k$ sentence queries for $10.5k$ videos of $8$ scenario domains, \ie{} Street View~(indoor), Street View~(outdoor), Game, Sports, Driving, Activity, TV Show, and Cooking.
The proposed \textvr requires one unified cross-modal model to recognize and comprehend texts, relate them to the visual context, and decide what text semantic information is vital for the video retrieval task.

Besides, we present a detailed analysis of \textvr in comparison to the existing datasets and design a novel multimodal video retrieval baseline for the text-based video retrieval task.
The dataset analysis and extensive experiments show that our \textvr benchmark provides many new technical challenges and insights over previous datasets for the video-and-language community.
The project website and github repo can be found at \href{https://sites.google.com/view/loveucvpr23/guest-track}{\color{blue}{$\tt CVPR 23 LOVEU$}} and \href{https://github.com/callsys/TextVR}{\color{blue}{$\tt TextVR$}}, respectively.
\end{abstract}


\begin{figure}[t]
\begin{center}
\includegraphics[width=0.48\textwidth]{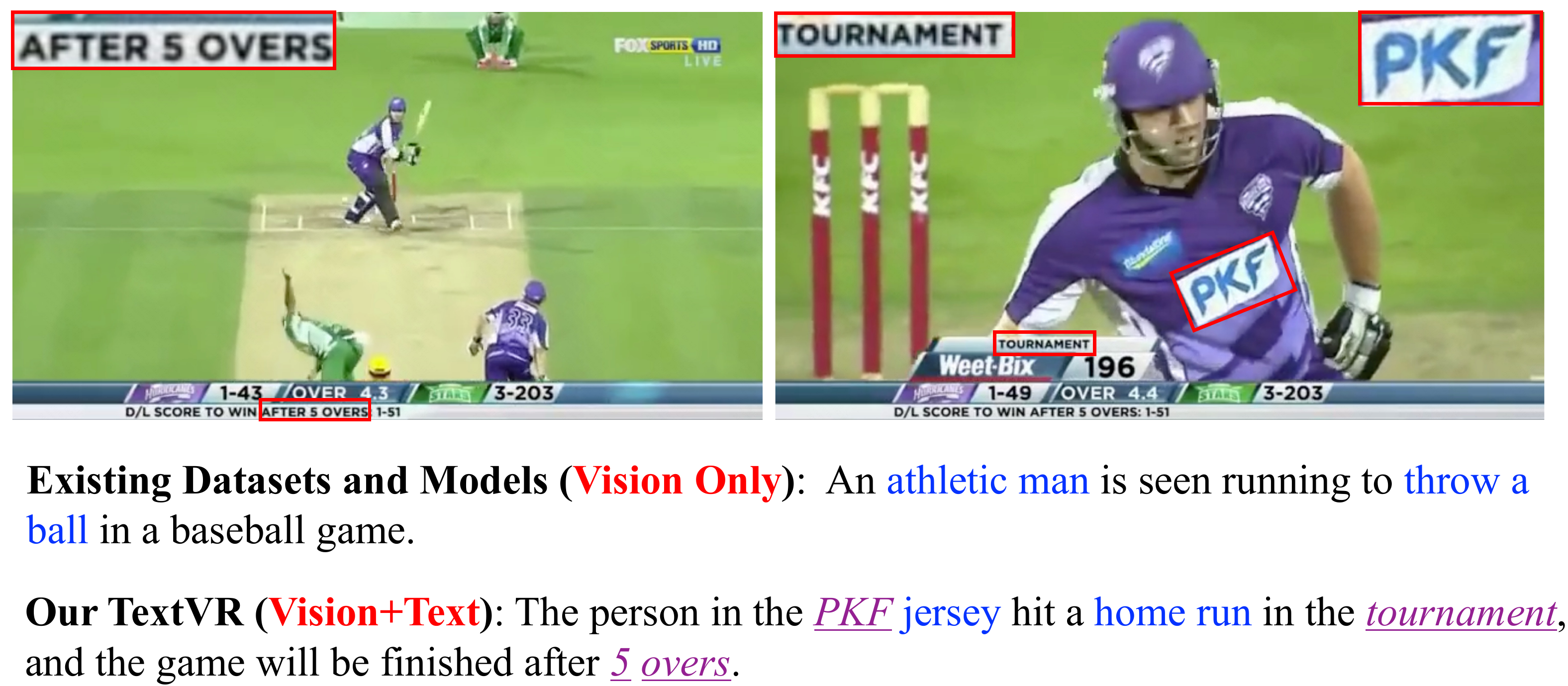}
\caption{\textbf{Existing video retrieval models cannot read, but \textvr can.} \textvr challenges a model to retrieve video with \textit{Vision} and \textit{Text} aggregation, which requires text reading comprehension, and visual reasoning. The video is from ActivityNet~\cite{krishna2017dense}. \textcolor[RGB]{0,0,255}{In blue} and \textcolor[RGB]{160,32,240}{purple} refer to vision and text information, respectively.
}
\label{fig2}
\end{center}
\end{figure}

\begin{figure*}[!t]
\begin{center}

\includegraphics[width=1\textwidth]{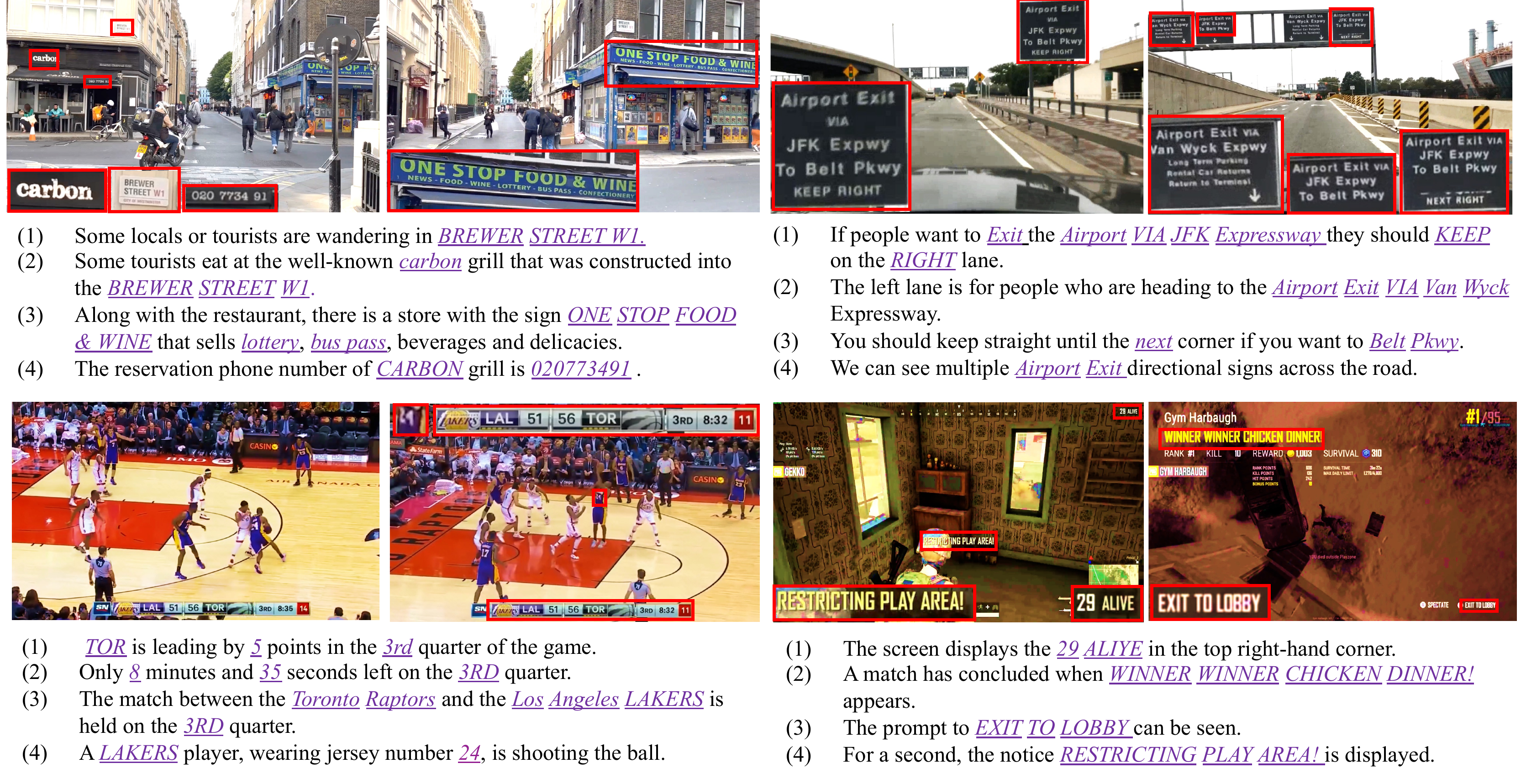}
\caption{\textbf{Illustration of \textvr}. \textcolor[RGB]{160,32,240}{In purple} and \underline{\textit{underlined}} are used to highlight the text information~(OCR) from video.}
\label{architecture}
\end{center}
\end{figure*}
\section{Introduction}
Cross-modal language-to-video retrieval~(VR)~\cite{kiros2014unifying,anne2017localizing,xu2016msr,wang2019vatex,caba2015activitynet,zhu2020actbert,wu2020synthetic} task is the foundation of the video-and-language task, which connects computer vision and natural language processing (NLP), and has attracted rapidly growing attention from both communities.
And the task aims to search for the most relevant video or language with the interrelated text content of sentence query and visual representation from video.
Recently, most works~\cite{wang2022object,luo2021clip4clip,bain2021frozen,cheng2021improving} focus on better visual representation and language-and-video alignment strategies, which largely promote the development of the field.
ClipBERT~\cite{lei2021less} adopts sparsely sampled clips to replace densely extracted video features, where only a few sparsely sampled short clips from a video are used at each training step.
Frozen~\cite{bain2021frozen} propose to train image and video simultaneously, where treating images as ‘frozen’ snapshots of the video, and gradually learning to attend to  increasing temporal context when trained on video datasets. %
Although the existing cross-modal language-to-video retrieval task has made extraordinary progress, it still can not learn some fine-grained semantics representation, \eg{} reading text in the video, which includes abundant and discriminative semantic representation.

Text, in human environments, plays a vital role in representing abundant semantics knowledge, which is indispensable for understanding video.
As shown in Fig.~\ref{fig2}, the information `\textit{athletic man}' and `\textit{throw a ball}' can be obtained from visual representation, but it is enough for understanding the video completely.
In real life, for human, the semantics `\textit{PKF athlete}' and `\textit{5 overs}' is the same important.
People always want to know which team is leading or what is the score now. 
\textit{Sometime, without text, humans even can not understand what happened!}
But the existing video retrieval models can not read and comprehend texts in the video, losing vital semantic information.
Therefore, how to enable and evaluate the reading comprehension ability of the video retrieval model still remain a huge challenge.

Existing benchmarks can not evaluate the reading ability of the model effectively, while they suffer from some limitations: 1) Without considering text semantics, their queries only focus on the information from visual representation. 
2) With the low proportion of videos that contain text, these datasets can not evaluate the effectiveness of text semantics for video-and-language retrieval tasks.
3) The video resolution of the existing dataset is not enough to read the fine-grained text, while most text regions in the video are blur and out-of-focus.
To break the dilemma, in this paper, we present a large cross-modal language-and-video retrieval dataset with reading comprehension, which considers the text semantics and visual entities~(\eg{} objects) into the sentence query.
Correspondingly, the proposed \textvr has three advantages.
\textbf{Firstly}, as shown in Fig.~\ref{architecture}, \textvr provides $42.2k$ sentence queries, which includes semantic information from text/OCR tokens~\footnote{“(Text Reading)~OCR tokens” in the paper refer to the text in the video, where one token is typically a word or phrase.}\\~(\eg{} `\textit{CARBON}', `\textit{020773491}') and visual context~(\eg{} `\textit{Tourist}', `\textit{restaurant}') simultaneously.
\textbf{Secondly}, $10.5k$ videos are collected from \textit{YouTube}\footnote{https://www.youtube.com/} and existing video retrieval datasets~(\eg{} ActivityNet~\cite{caba2015activitynet}), each video on the dataset includes at least $10$ text instances. 
\textbf{Thirdly}, videos on \textvr present high resolution, which enables more accurate text/OCR tokens from the text reading model.
Besides, the intuitional advantages, \textvr also \\ presents 
some new insights and technical challenges: 
1) Which ones are \textbf{relevant}, while there exist up to hundreds of texts/OCR tokens?
Dealing with hundreds of texts is extremely \textit{time-consuming}, and universal semantics from irrelevant texts/OCR tokens even cause \textit{negative impact}, \eg{} the word `\textit{NBA}'~(Fig.~\ref{architecture}) in NBA video series represents confusing and common semantics.
2) How to \textbf{fuse} semantic representation and visual representation effectively.

To explore and evaluate the benchmark, we also present a simple yet effective \textit{\textbf{s}cene \textbf{t}ext \textbf{a}w\textbf{a}re} cross-modal video retrieval model~(StarVR), as the baseline.
Inspired by the previous dual-encoder methods~\cite{luo2021clip4clip,cheng2022vista}, we propose a transformer-based cross-modal video retrieval model, which fuse vision and scene text representation in one feature embedding. 

Our main contributions are summarized as follows:
\begin{itemize}
    \item We propose a large cross-modal video retrieval dataset with text reading comprehension, named \textvr, which includes $42.2k$ sentence queries for $10.5k$ videos of $8$ scenario domains.
    Different from previous benchmarks, \textvr requires models to retrieve video with both of semantic information from \textbf{text/OCR tokens} and \textbf{visual context} simultaneously.

    \item For cross-modal video retrieval task on \textvr, we present \textbf{sufficient experiment analysis}~(\eg{} unique evaluation for reading comprehension ability), \textbf{new insights} and \textbf{new challenges}~(\eg{} negative impact from irrelevant and noise and text/OCR tokens).

    \item An \textit{\textbf{S}cene \textbf{T}ext \textbf{A}w\textbf{a}re} \textbf{V}ideo \textbf{R}etrieval baseline, \starvr, fusing both \textbf{semantics} from reading comprehension and visual representation is provided for the new task. Experiments show that current state-of-the-art cross-modal video retrieval methods fail on \textvr, while our \starvr with scene text semantic representation, gets encouraging results.
    
\end{itemize}

\section{Related work}

\subsection{Text-based Image Retrieval}
\textbf{Method.}
There are mainly two branches for text-based image retrieval: retrieval with word-level query~\cite{mishra2013image,ghosh2015efficient,wang2021scene,miyawaki2022scene} and retrieval with sentence-level query~\cite{cheng2022vista,mafla2021stacmr}.
For the former, Mishraet \textit{et al.}~\cite{mishra2013image} proposed to detect approximate locations of characters, and then impose spatial constraints to generate a ranked list of images.
Mafla \textit{et al.}~\cite{MAFLA2021107656} adopted the nearest neighbor search of the textual representation of a query over the outputs of a single shot text detection architecture to retrieve all images containing the queried texts.
Liang.~\cite{LIANG20124225} proposed to retrieve words in handwritten documents using a character-based modeling approach and synthesized words.
Ghosh \textit{et al.}~\cite{ghosh2015efficient} introduced Pyramidal Histogram of Characters (PHOC)~\cite{almazan2014word} to compare text regions with the query word and designs a space-based PHOC to determine similar attribute.
TDSL~\cite{wang2021scene} proposed an end-to-end trainable retrieval network, which jointly optimizes the procedures of scene text detection and cross-modal similarity learning.
For retrieval with sentence query, Stacmr~\cite{mafla2021stacmr} proposed a scene-text aware cross-modal retrieval method, which uses specialized semantic representations from captions and text from the visual scene.
%
VisTA~\cite{cheng2022vista} adopted transformer blocks to directly encode image patches
%
and fuse scene text embedding to learn an aggregated representation for cross-modal retrieval. 

\begin{table*}[t]
    \centering
	\caption{\textbf{Statistical Comparison with existing benchmarks.} `Text' denotes any input text feature from OCR, \ie{} subtitle, title scene text, and others. `Subtitle' refers to the provided subtitle information without OCR. `k' and `m' denote `thousand' and `million'. `Resolution' refers to the average resolution for all frames.}
	\label{table1}
	\input{table/table1}

\end{table*}

\textbf{Benchmark.}
There are mainly three common datasets for text-based image retrieval.
As shown in Table.~\ref{table1}, Street View Text dataset (SVT)~\cite{wang2011end} is collected from Google Street View, including 349 images and 427 annotated query words.
IIIT Scene Text Retrieval dataset\\
~(STR)~\cite{mishra2013image} consists of 50 query words and 10,000 images, focusing on variations of fonts, styles, and viewpoints.
And Coco-Text Retrieval dataset~(CTR) is sampled from Coco-Text~\cite{veit2016coco}, including 500 query words and 7,196  images.
COCO-Text Cap~\cite{mafla2021stacmr} firstly proposed a scene text aware sentence-level query for retrieval task, which includes 53.0k sentence query for 10.6k images.
The above text-based retrieval benchmarks focus on word-level queries, which is not matched with general retrieval task with sentence-level query.
No one of the above benchmarks tries to explore video-level retrieval with text reading, while cross-modal video retrieval has become an active research area. 
To promote video retrieval, we create a large video retrieval dataset with reading comprehension.

\subsection{Cross-Modal Video Retrieval}
\textbf{Method.} Various methods~\cite{lei2021less,bain2021frozen,zhu2020actbert,wang2022object} based on deep learning have been proposed and have improved the performance considerably for video-and-language cross-modal retrieval.
Beltran~\textit{et al.}~\cite{FERNANDEZBELTRAN201672} presented to perform the video retrieval task and cope with the semantic gap challenge by means of latent.
ClipBERT~\cite{lei2021less} proposed end-to-end learning for retrieval tasks, which employs sparse sampling to replace the densely sampled clips.
VSRNet~\cite{SUN2021108027} proposed a text-aligned attention mechanism to efficiently generate a temporal proposal and a collaborative ranking strategy to improve the performance of video retrieval.
Chiang \textit{et al.}~\cite{CHIANG2022108807} retrieve videos by adopting ConvLSTM to extract spatial and temporal characteristics from video and uses an encoder-decoder structure to learn multi-scale video embedding to compare the similarity of each video pair in terms of both spatial and temporal characteristics.
Frozen~\cite{bain2021frozen} takes advantage of both large-scale image and video datasets, where the model is flexible and can be trained on both image and video text datasets.
OA-transformer~\cite{wang2022object} presented an object-centric approach, which leverages the bounding
boxes and object tags to guide the training process.
The existing cross-modal retrieval models all focus on visual representation, while ignore the strong semantic representation from scene text.

\textbf{Benchmark.} As shown in Table.~\ref{table1}, the existing benchmarks~\cite{xu2016msr,chen2011collecting,anne2017localizing,wang2019vatex,caba2015activitynet} have dramatically improved the development of video-and-language retrieval task.
MSR-VTT~\cite{xu2016msr}, as the most popular benchmark, including $10K$ YouTube videos with $200K$ descriptions, and each video clip is annotated with 20 English sentences by Amazon Mechanical Turks.
MSVD~\cite{chen2011collecting} dataset consists of about $120K$ sentences for $2k$ video collected during the summer of 2010. 
VATEX-EN-R~\cite{wang2019vatex} is a large multilingual video description dataset, which contains over $35k$ videos and $349.9k$ captions covering $600$ human activities. 
ActivityNet~\cite{caba2015activitynet} contains $200$ different types of activities and a total of $849$ hours~($100k$ videos and $100k$ captions) of videos collected from YouTube.
The existing cross-modal video retrieval benchmarks suffer from several limitations, including ignoring text semantics and low resolution of the video.
To overcome these limitations, we present a cross-modal video retrieval dataset with reading comprehension.

\subsection{Video OCR System}
In recent years, video text spotting~\cite{yin2016text} have made great progress thanks to several video text spotting models~\cite{wu2022end,wu2021bilingual,Google_api,wang2017end,wu2022real,wu2020texts,wu2023icdar}.
Wang \textit{et al.}~\cite{wang2017end} tracks text by associating texts in the current frame and several previous frames to obtain final results. 
Wu \textit{et al.}~\cite{wu2022end} propose to track and recognize text with the inherited query from the previous frame.
TransDETR~\cite{wu2022end} proposed an end-to-end trainable video text spotting framework with Transformer, which simultaneously solves text detection, tracking, and recognition tasks in one framework with one concept, \ie{} text query.
In this paper, we adopt TransDETR~\cite{wu2022end}, Kuaishou VideoOCR api~\cite{KuaiShou_api}, Google VideoOCR api~\cite{Google_api} as the base video ocr models to obtain the text/OCR tokens.

\section{\textvr Benchmark}

\subsection{Data Collection and Annotation}
\textbf{Data Collection.} To obtain abundant videos data, we collect videos from three parts: video retrieval dataset~(\eg{} ActivityNet~\cite{caba2015activitynet}, YouCook2~\cite{zhou2018towards}), video OCR datasets~(\eg{} RoadText-1k~\cite{reddy2020roadtext}), and \textit{YouTube}\footnote{https://www.youtube.com/}. 
For video retrieval datasets, we first obtain the OCR tokens of the videos with Google VideoOCR API~\cite{Google_api} and collect those videos that contain at least $5$ text instances.
For video OCR datasets, we directly select videos with the ground truth of OCR.
Considering the requirement of high resolution and the scenario distribution, ActivityNet~\cite{caba2015activitynet}, YouCook2~\cite{zhou2018towards}, and RoadText-1k~\cite{reddy2020roadtext} are used as the video source representing three scenario class, \ie{} \textit{Activity, Cooking, and Driving}.
And we select $3.151k$, $602$, and $657$ videos from the three datasets, respectively.
Besides, other $6,090$ videos, for five scenario, \ie{}`Sports', `Game', `Street View Indoor', `Street View Outdoor', and `News\&Movie' are collected from YouTube.
As a result, we collect $10.5k$ videos, where including at least $5$ text instances per video.

\textbf{Annotation.}
We invite a professional annotation team from Appen\footnote{https://appen.com/}, where each annotator is a native speaker of English.
Similar to TextCaps~\cite{sidorov2020textcaps}, annotators are asked to describe an around $15$ seconds video in one sentence which would require \textit{reading texts} in the video.
And the detailed requirements for each description are as follows:
\begin{itemize}
    \item The sentence description requires \textit{reading the text} in the video and relating them with visual context, \eg{} objects, color, shape.

    \item The sentence should contain at least 8 words.
    
    \item
    Each description should be a single sentence, not a combination of multiple short sentences.
    
    \item The sentence should be grammatically correct.
    
    \item No profanity or any other form of objectionable text should be used.
    
    \item Well-known brand~(\eg{} BBC, NBA) can not be used as the only text in the description. 
\end{itemize}
After finishing the annotation of videos, there are two rounds of verification for the annotation, one by Appen, and one by us. 
Ourselves, five evaluators are asked to check the annotations, which are approved if the accuracy is greater than $95\%$.
And the annotator will be replaced if the accuracy is under $80\%$.
Around $40\%$ annotations are rejected during the whole labeling period.
The description will be rejected if any one of the above requirements is violated.
Finally, we annotate $42.2k$ sentence descriptions for $10.5k$ videos, where each video clip is annotated with about $4$ sentences.
And we divide the dataset into two parts: the training set with $29.5k$ sentence descriptions for $7.1k$ videos, and the testing set with $12.7k$ sentence descriptions for $3.4k$ videos.
As a labor-intensive job, the whole labeling process takes $30$ men in one week, \ie{} $1,680$ man-hours.

\subsection{Dataset Comparison and Analysis}
The basic comparison between \textvr and previous datasets is summarized in Table~\ref{table1}. 
Besides, to highlight the feature of \textvr, \eg{} \textit{the sentence query contains the textual information present in the video}, we compare \textvr with other prominent video retrieval datasets~(\eg{} MSR-VTT~\cite{xu2016msr}, ActivityNet~\cite{caba2015activitynet}) in three aspects: \ie{} video resolution, number of words per sentence query, number of text/OCR tokens per sentence query.

\textbf{Comparison for basic information}. Table.~\ref{table1} \\
presents the detailed comparison of basic information~(\eg{} video, query) between \textvr and text-based image retrieval datasets and general video retrieval datasets.
We want to highlight three points: 1) Different from the previous datasets, \textvr, as the first video retrieval dataset, considers scene text semantics in the video, relating them with visual context. 
YouCook2 tackles cross-modal retrieval task with subtitle ground truth, which is not the original text without text reading~(\ie{} Video OCR).
In another word, the subtitles are added artificially, which does not belong to the scene text.
2) Videos in \textvr present higher resolution~(\ie{} average 1080p per video), which is helpful for the fine-grained representation extraction, \ie{} video text in this paper.
3) Abundant video scenarios are provided on \textvr, including some new scenario, \eg{} \textit{Game}.

\begin{figure}[t]
\begin{center}

\includegraphics[width=0.48\textwidth]{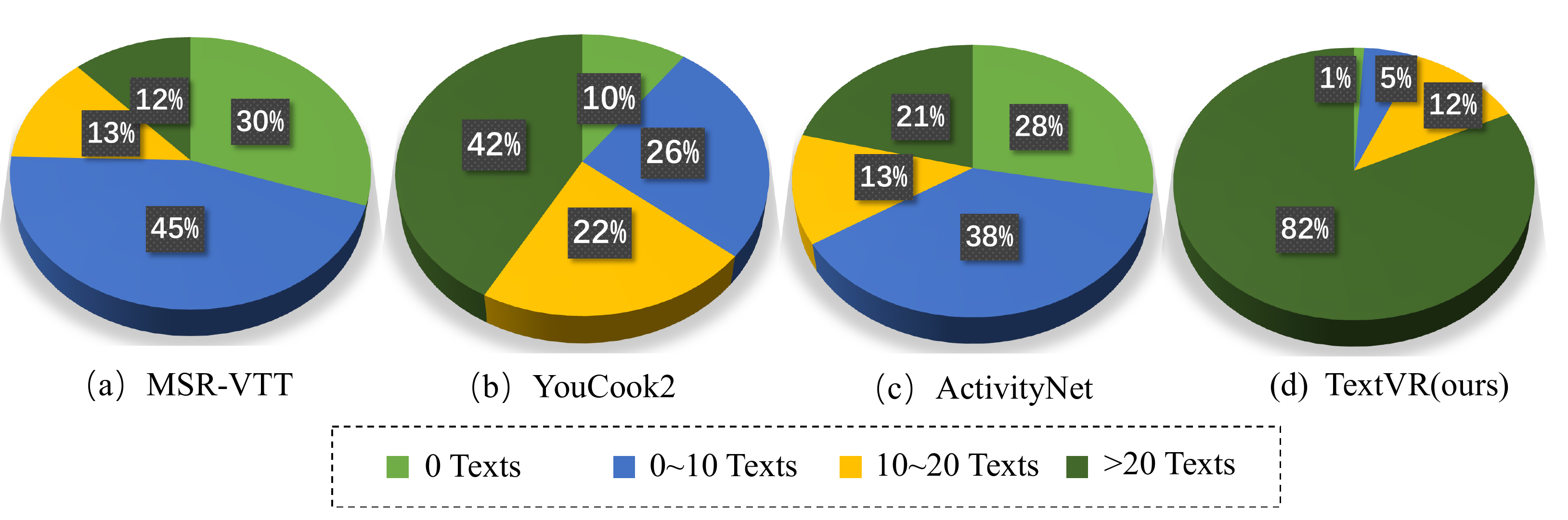}
\caption{\textbf{Number of text/OCR tokens per video.} "\%" denotes the percentage of the corresponding video number over the whole data. Google VideoOCR~\cite{Google_api} is used to obtain the number of text/OCR tokens at each video.}
\label{Statictical_Analysis}
\end{center}
\end{figure}

\textbf{Number of text/OCR tokens per video}. As shown in Fig.~\ref{Statictical_Analysis}, we present the percentage of videos with different text/OCR token numbers. 
$30\%$ videos on MSR-VTT do not includes text, and $45\%$ videos only contains less than $10$ text instances.
By comparison, videos on YouCook2~\cite{zhou2018towards} contain more texts, $45\%$ videos have at least $20$ text instances. 
In summary, videos on our \textvr contain more texts, which represent more abundant text semantics.
$82\%$ videos on \textvr contains at least $20$ text instances.

\begin{figure}[t]
\begin{center}
\includegraphics[width=0.48\textwidth]{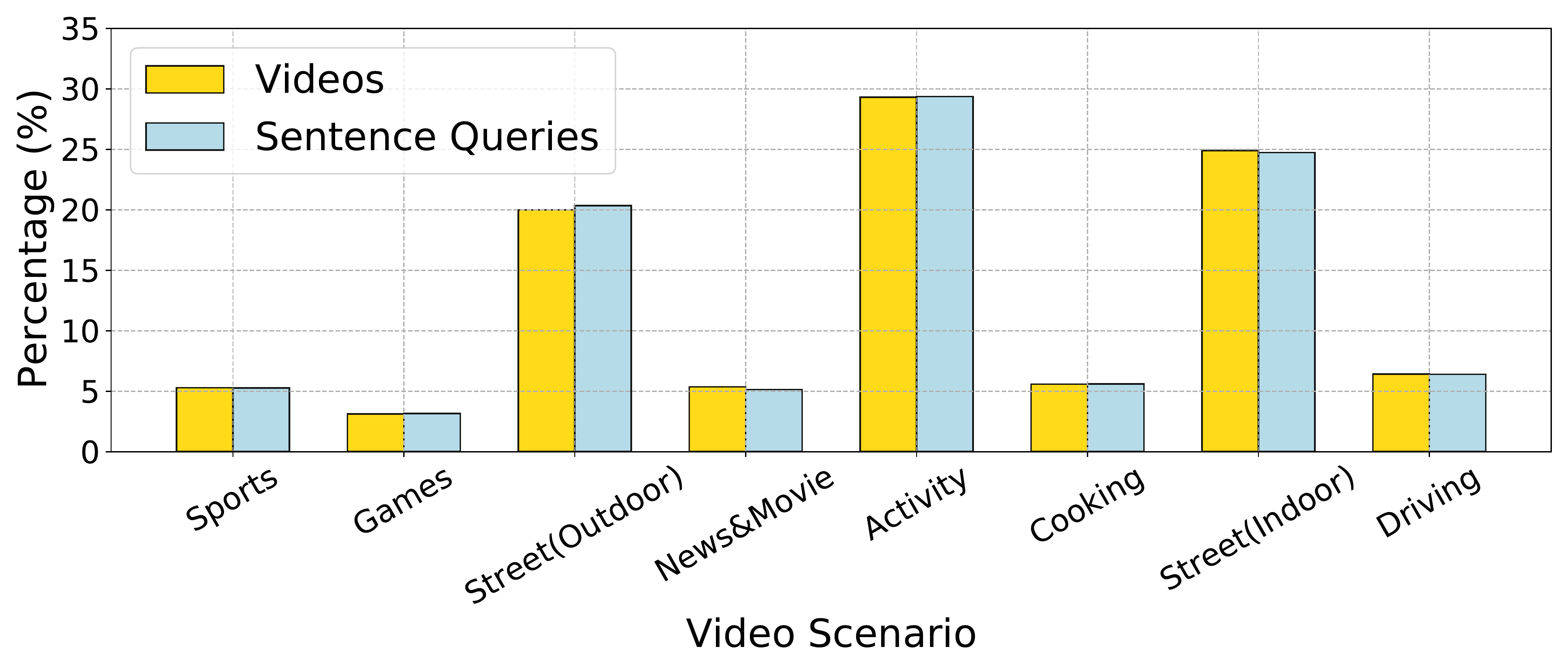}
\caption{\textbf{Data Distribution for $8$ Open Scenarios.} "\%" denotes the percentage of each scenario data over the whole data.}
\label{Scenarios}
\end{center}
\end{figure}

\begin{figure}[t]
\begin{center}
\includegraphics[width=0.48\textwidth]{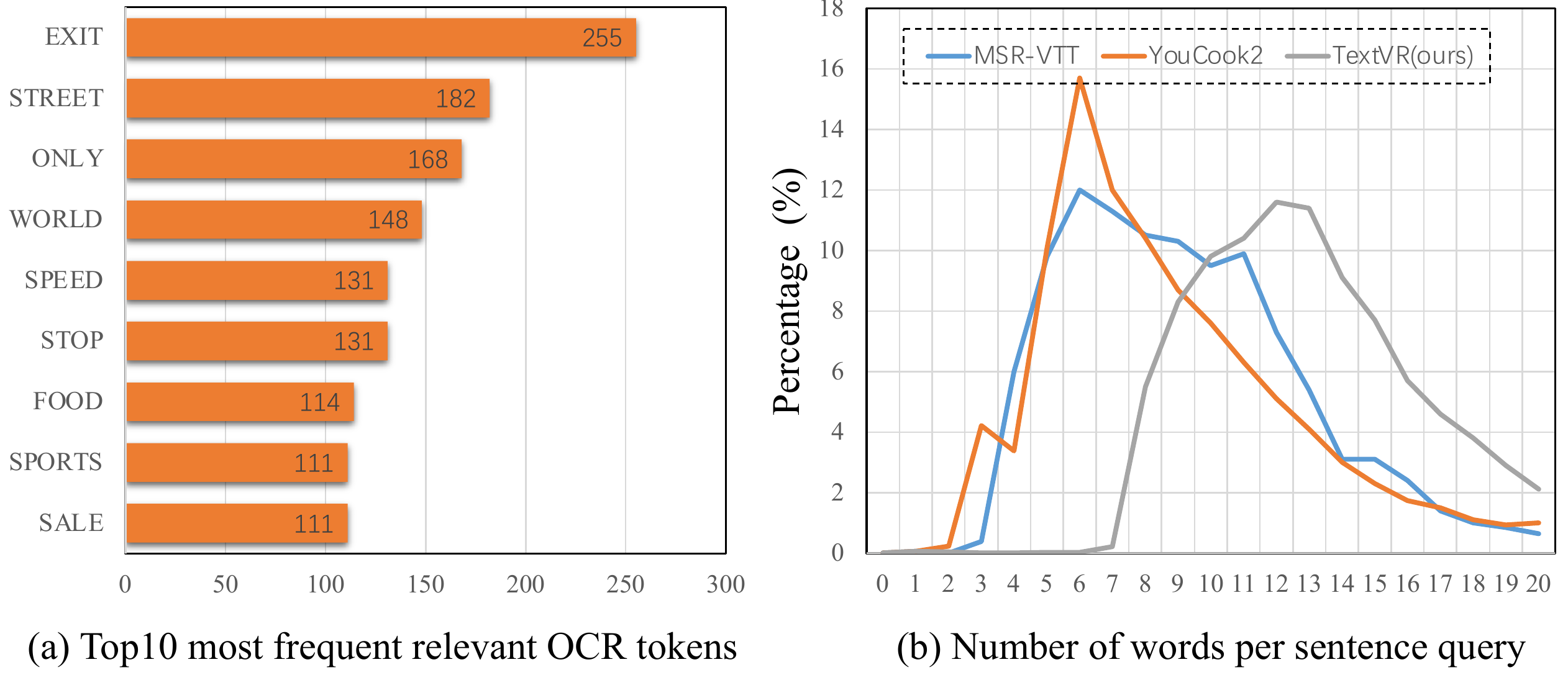}
\caption{\textbf{Top 10 most frequent relevant text/OCR tokens and Number of words per query.} Google VideoOCR~\cite{Google_api} is used to obtain the OCR tokens.}
\label{top10}
\end{center}
\end{figure}

\begin{figure*}[!t]
\begin{center}
\includegraphics[width=1\textwidth]{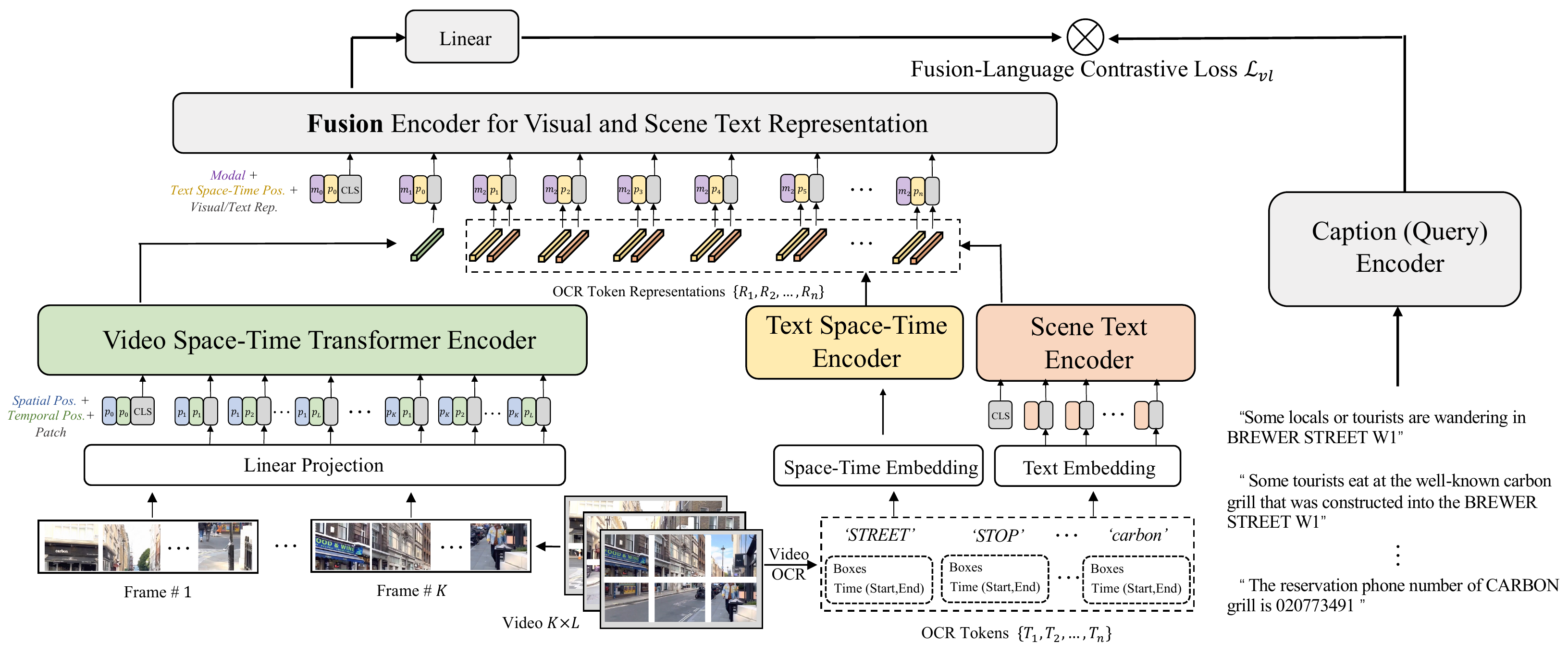}
\caption{\textbf{Framework of \starvr}. The whole pipeline includes four parts: a Space-time transformer encoder for visual feature extraction, a space-time and scene text encoder for text/OCR tokens feature extraction, a fusion encoder, and a caption encoder for query sentences. }
\label{baseline}
\end{center}
\end{figure*}

\begin{figure}[t]
\begin{center}
\includegraphics[width=0.48\textwidth]{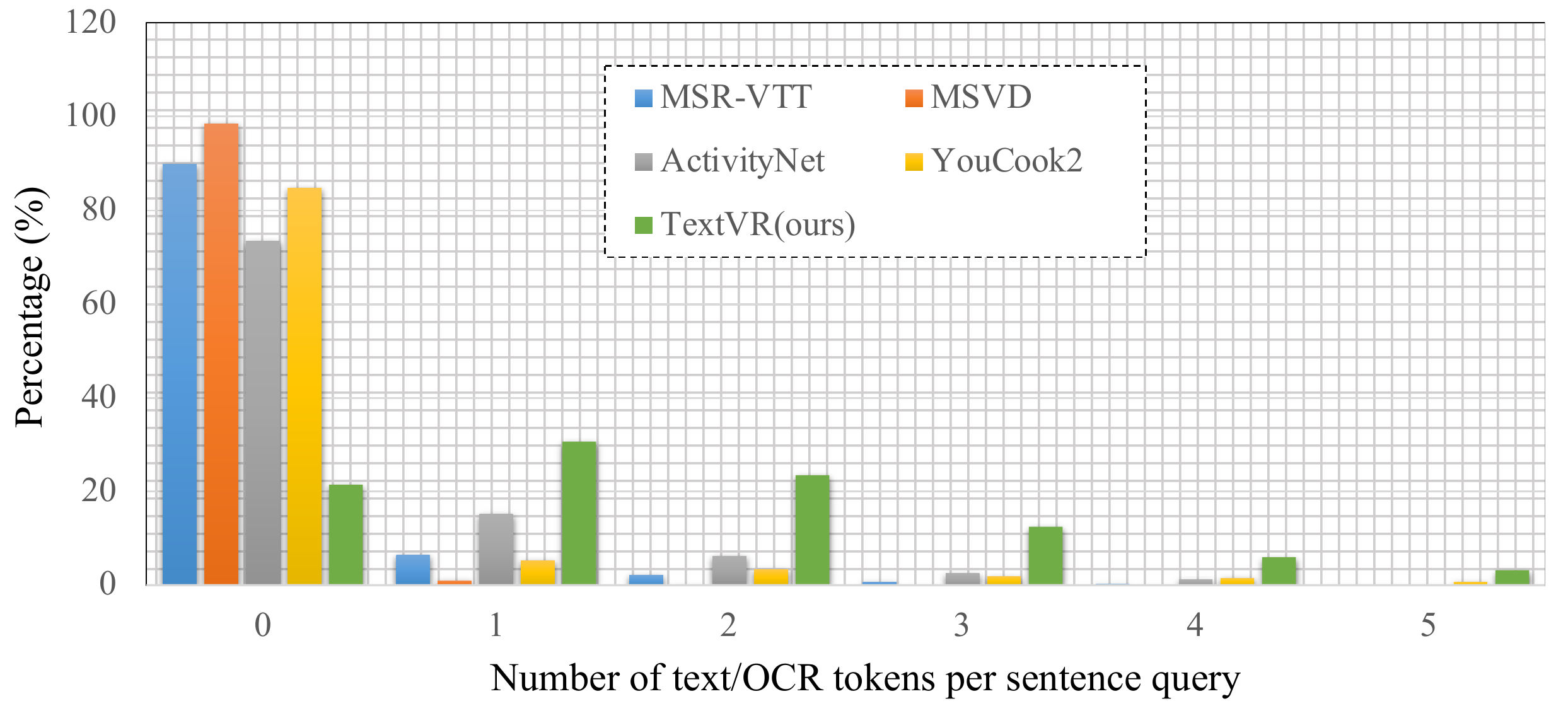}
\caption{\textbf{Distribution of OCR tokens in sentence query for different datasets.} Google VideoOCR~\cite{Google_api} is used to obtain the OCR tokens, then we match them with sentence queries by judging whether the OCR token appears in the sentence.}
\label{text_per_caption}
\end{center}
\end{figure}

\textbf{Distribution for 8 open scenarios}. 
Fig.~\ref{Scenarios} shows the distribution of $8$ video scenarios.
The distribution of sentence query is similar to that of video, while the video is annotated by four sentences.
$Activity$ scenario occupy the largest percentage~(\ie{} around $29\%$) of videos, while only around $3\%$ videos belong to $Games$ class.

\textbf{Top 10 most frequent relevant text/OCR tokens}. 
We also present the Top 10 most frequent Video OCR tokens in Fig.~\ref{top10}~(a).
The most frequent OCR token is the \textit{EXIT}, which always appears in \textit{Street View} scenario.
Actually, the Top 10 most frequent OCR tokens are not a good choice to be used in the description, while they represent the same semantics in hundreds of videos simultaneously.

\textbf{Number of words per sentence query}. As shown in Fig.~\ref{top10}~(b)
The average sentence query length is around $7$ words for MSR-VTT~\cite{xu2016msr}, $8$ words for YouCook2~\cite{zhou2018towards}, respectively. 
And the average length for \textvr is around $12.5$, obviously larger than the previous benchmarks. 
The main cause is that the sentence query of \textvr includes at least $8$ words, while we argue the detailed description is necessary for containing both scene text and visual context.

\textbf{Number of text/OCR tokens per sentence query}. 
Fig.~\ref{text_per_caption} compares the percentage of sentence queries with a particular number of OCR tokens for different datasets.
Most sentence queries on \textvr all contain at least one text/OCR token from the video, while the queries of previous datasets~(\eg{} MSR-VTT, YouCook2) almost do not contain the text/OCR token from the corresponding video.
This is reasonable and explainable with two points: most videos on previous datasets contain little texts~(less than $10$) in Fig.~\ref{Statictical_Analysis}, and the corresponding descriptions only focus on visual context, \ie{} \textit{human activity}, without text semantics.

\section{Baseline: Scene Text Aware Video Retrieval}
As shown in Fig.~\ref{baseline}, a dual-encoder transformer architecture is designed for baseline, \ie{} \starvr,
, which learns a discriminative cross-modal representation with contrastive losses between fusion-and-language pairs.

\subsection{Video Encoder}
The input video is divided into $K \times L$ space-time patches, whose  size is $P \times P$.
The patches $\boldsymbol x \in \mathbb{R}^{K \times L \times 3 \times P \times P}$ is flattened with one layer CNN, forming a sequence of embeddings $\boldsymbol z\in \mathrm{R}^{KL \times D}$, where $D$ refers to the kernel number of the convolutional layer.
Similar to Frozen~\cite{bain2021frozen}, after adding temporal and spatial positional embeddings, the input tokens are fed into a stack of space-time transformer blocks to obtain the final vision representation $\mathbf{V}_{K,L}$.

\subsection{Space-Time Encoder for Scene Text}
\label{OCR token}
A set of scene text instances $\{T_1,T_2,...,T_n\}$ is obtained from each video by an Video OCR model~\cite{wu2022end,Google_api}, where each text instance includes recognition result $\mathbf{o}^{word}$, bounding box $\mathbf{o}^{bbox}$, and time slot~(start $t_S$ and end $t_E$ frame) $\mathbf{o}^{time}$.
If there is no detected text, the set is an empty set $\emptyset$.
Inspired by positional embedding from VI-BERT~\cite{su2019vl}, for each OCR token $T_i$, we extract the text space-time representation $\mathbf{R}_{ST}$ by embedding box coordinates and time slot of the text instance.
The bounding boxes of each text trajectory are represented by an $8$-d vector, as $( \frac{x_{LT}}{W}, \frac{y_{LT}}{H}, \frac{x_{RB}}{W}, \frac{y_{RB}}{H}, \frac{x_E - x_S}{W}, \frac{y_E - y_S}{H}, \frac{w_E-w_S}{W}, \\\frac{h_E-h_S}{H})$, where $(x_{LT}, y_{LT})$ and $(x_{RB}, y_{RB})$ are the \textit{average value} of the top-left and bottom-right coordinates during tracking period respectively, and $W$,$H$ are the width and height of input frame.
$(x_{S}, y_{S})$ and $(x_{E}, y_{E})$ are the center coordinate of the start and end frames, respectively.
$(w_{S}, h_{S})$ and $(w_{E}, h_{E})$ are the width and height of the bounding box at start and end frame, respectively.
Each time slot of text is characterized by a $3$-d vector,
$( \frac{t_S}{K}, \frac{t_E}{K}, \frac{t_E}{K} - \frac{t_S}{K})$, where $K$ is the length of whole video frames.
Then the $8$-d vector from the box and $3$-d vector from the time slot are concatenated into an $11$-d vector.
Finally, for $n$ OCR tokens, the $n$ $11$-d vectors are embedded to text space-time representations $\mathbf{R}_{ST}$ ( $n\times 768$ in this paper) by one-layer perception.

\subsection{Scene Text and Query Encoder}
%
The scene text encoder is a multi-layer bidirectional transformer encoder, the same as the sentence query encoder.
For all texts in the same trajectory, the text recognition result with the highest confidence is used.
For the final text encoding, we use the [\texttt{EOS}] token output of the final layer.
Therefore, for $n$ OCR tokens, we can obtain $n$ scene text context representation $\mathbf{R}_{TC}$ (size: $n\times 768$ in this paper)
For query encoder, similarly, the activations at the [\texttt{EOS}] token are treated as the feature representation of the sentence query $\mathbf{w}_j$, where $j$ denotes $j$-th sentence.


\subsection{Fusion Encoder for Visual and Scene Text Representation}
With vision space-time representation $\mathbf{V}_{K,L}$, text space-time representation $\mathbf{R}_{ST}$, and scene text context representation $\mathbf{R}_{TC}$, the final fusion embedding can be calculated as following:
\begin{equation}
\begin{aligned}
    &\mathbf{R} \gets \mathrm{MHSA}([\mathbf{V}_{K,L}; \mathbf{R}_{ST} + \mathbf{R}_{TC})
\end{aligned}
\end{equation}
where $\mathrm{MHSA}(\cdot)$ denotes the multi-head self-attention layer.

\subsection{Cross-Modal Contrastive Learning}
Following previous retrieval setting~\cite{bain2021frozen,wang2022object,lei2021less,luo2021clip4clip}, matched language-video pairs in the batch are treated as positives, and all other pairwise are treated as negatives. We minimize the sum of two losses, \ie{} video-to-language and language-to-video:
\begin{equation}
\label{eq:loss1}
    \mathcal{L}_{v2l} = -\frac{1}{N}\sum_i^N\log{\frac{\exp(x_i^\top y_i / \sigma)}{\sum_{j=1}^{N} \exp(x_i^\top y_j / \sigma)}}
\end{equation}
\begin{equation}
\label{eq:loss2}
    \mathcal{L}_{l2v} = -\frac{1}{N}\sum_i^N\log{\frac{\exp(y_i^\top x_i / \sigma)}{\sum_{j=1}^{N} \exp(y_i^\top x_j / \sigma)}}
\end{equation}
where $x_i$ and $y_j$ are the normalized fusion embedding of $i$-th video and the $j$-th sentence query $\mathbf{w}_j$ respectively in a batch of size $N$ and $\sigma$ is the temperature.

\section{Experiments}
\subsection{Implementation Details}
All the experiments are conducted on PyTorch with Tesla V100 GPUs. We use Frozen~\cite{bain2021frozen} as our basic network. 
AdamW~\cite{loshchilov2017decoupled} as the optimizer for total 50 epochs with the initial learning rate of 3e\mbox{-}5. The learning rate decays to 3e\mbox{-}6 at 30 epochs.
For a fair comparison, we do not use any extra dataset to pre-train the network in our experiment.
To provide comprehensive analysis, we adopt and compare the OCR tokens from three Video OCR models, \ie{} TransDETR~\cite{wu2022end}, Google Video OCR API~\cite{Google_api}, and Kuaishou Video OCR API~\cite{KuaiShou_api}.
TransDETR~\cite{wu2022end} is an open-source, end-to-end video ocr model, where the weight trained on ICDAR2015 video~\cite{karatzas2015icdar} is used in this paper.
\textbf{Evaluation Metric}
Following previous video retrieval benchmarks~\cite{xu2016msr,luo2021clip4clip}, we adopt the average recall at K(R@K), median rank~(MdR), and mean rank~(MnR) over all queries as the metric. 
We consider a prediction correct if the predicted video matches the ground-truth video.
Generally, the higher R@K and lower MdR, MnR show better performance.

\begin{table*}[t]
    \centering
    \small 
    \setlength{\tabcolsep}{2mm}
    \input{table/table3}
    \caption{\textbf{Impact of Text Semantic Representation for Different Dataset.} \green{In green} are the gaps compared to the vision-only baseline. MSR-VTT adopts ‘Training-9K’ setting~\cite{gabeur2020multi}. Obvious improvements from abundant text semantic representation on \textvr.}
    \label{tab:table3}
\end{table*}

\begin{table*}[t]
    \centering
    \small 
    \setlength{\tabcolsep}{2mm}
    \input{table/table5}
    \caption{\textbf{Ablation for \starvr and Cross Evaluation} . Trained on existing datasets can not obtain satisfactory performance on \textvr.}
    \label{tab:table5}
\end{table*}

\subsection{Analysis for \textvr}

\textbf{Impact of Text Semantic Representation for different Dataset.} 
The semantics of Text/OCR tokens, as the unique point of \textvr, is the core of the paper.
Table.~\ref{tab:table3} presents the impact of text semantics for different datasets.
MSR-VTT~\cite{xu2016msr} and YouCook2~\cite{zhou2018towards}, as the most popular and relevant video retrieval benchmarks respectively, are used to compare our \textvr.
With text semantic representations, MSR-VTT~\cite{xu2016msr} presents little to gain with $0.5$ R@1 and $0.3$ MdR.
Similarly, YouCook2~\cite{zhou2018towards} shows weak improvement for R@1, even negative impact for MdR, while its subtitle semantic representation brings huge improvement.
There are main two points for the results: 1) the subtitle is not the text in the video, which can not be detected by the video OCR model. 2) the ground truth of the subtitle is used without the loss of accuracy from the OCR model.
By comparison, \textvr still shows the effectiveness from scene text semantic presentation with $9.1$ R@1 and $13.0$ MdR improvements, while there existing recognition error from video ocr.
The experimental result proves the unique contribution of \textvr on evaluating the effectiveness of scene text semantic representation.

\begin{figure*}[!t]
\centering
\subfloat[]{\includegraphics[width=2.1in]{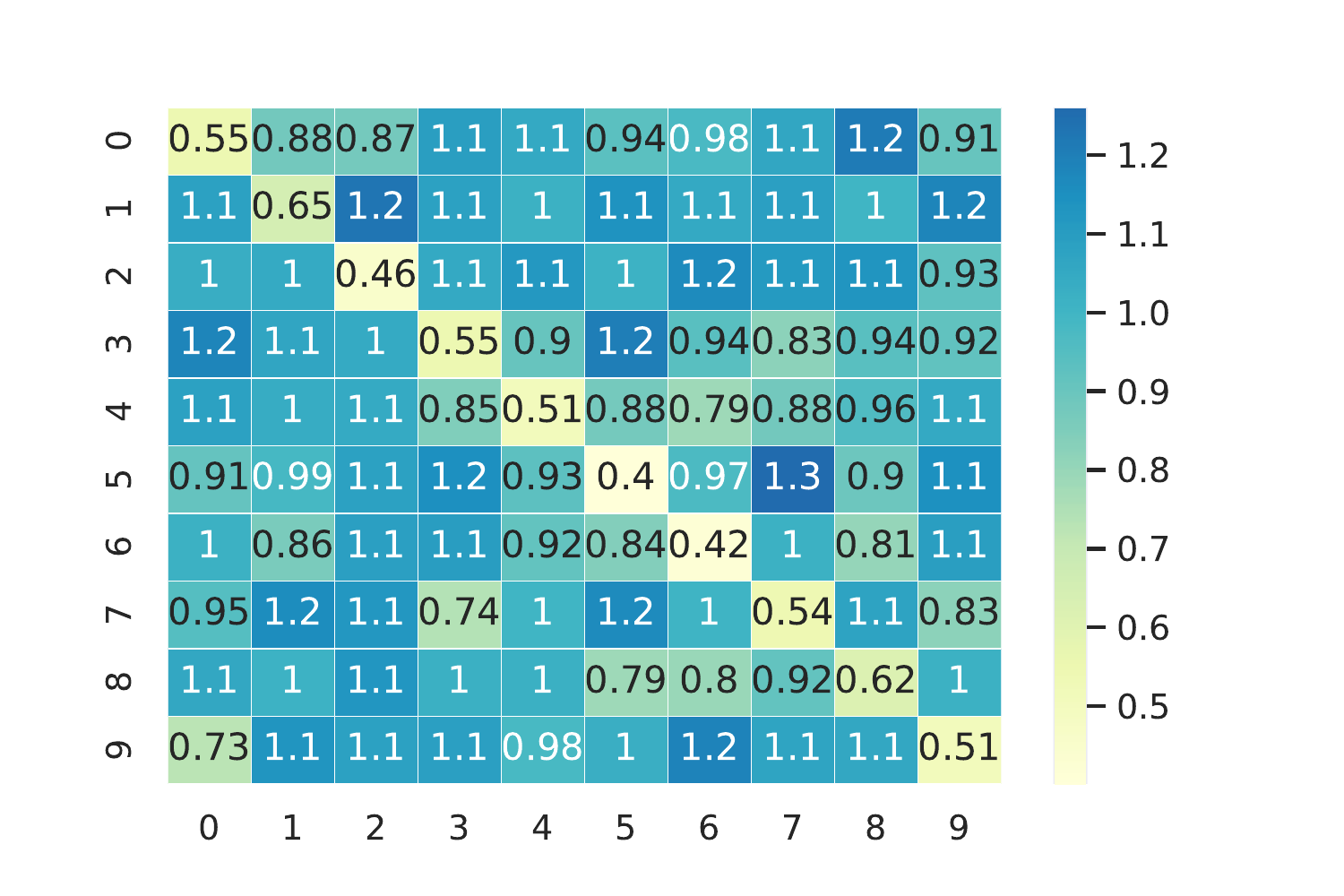}%
\label{fig:3a}}
\hfil
\subfloat[]{\includegraphics[width=2.1in]{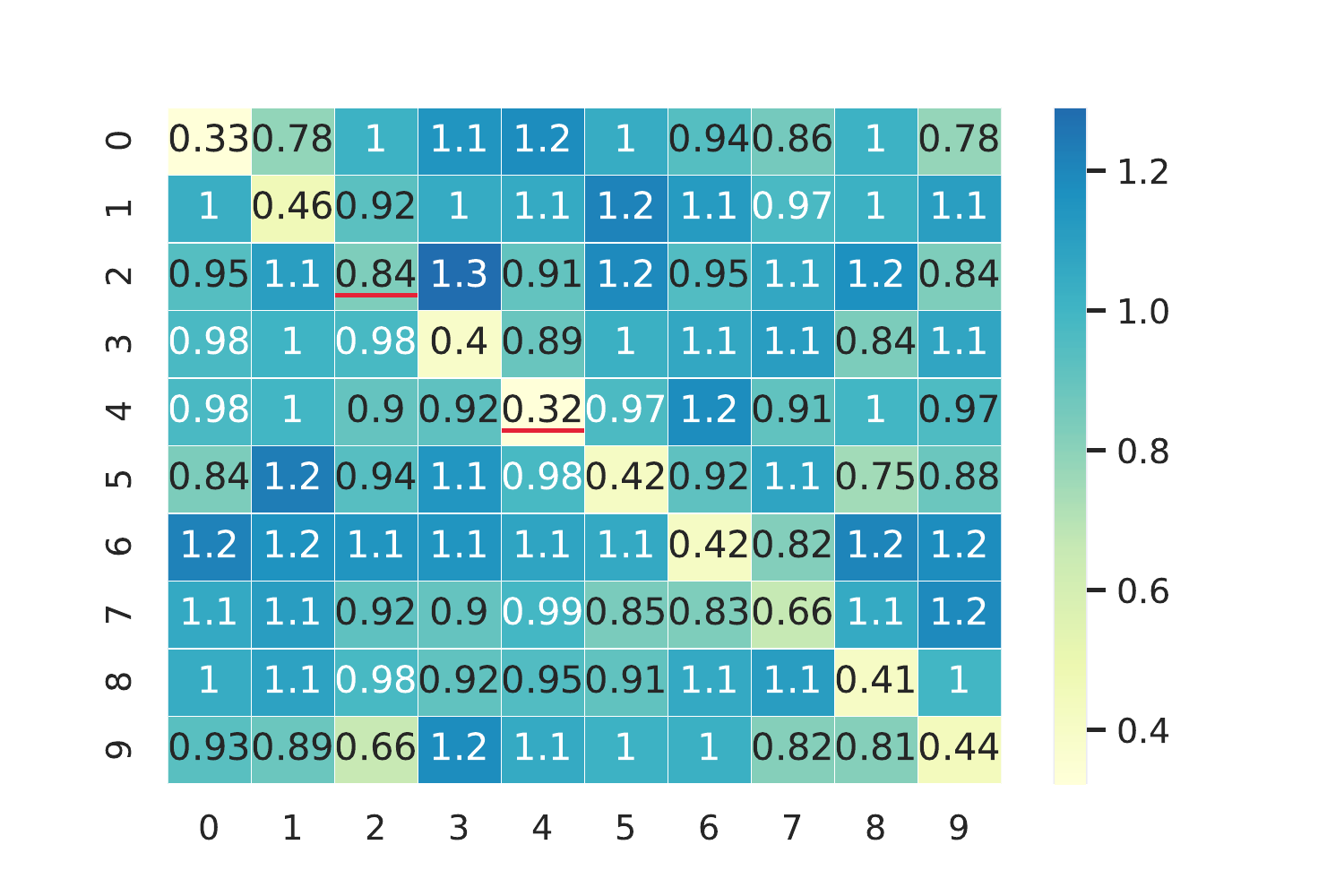}%
\label{fig:3b}}
\hfil
\subfloat[]{\includegraphics[width=2.3in]{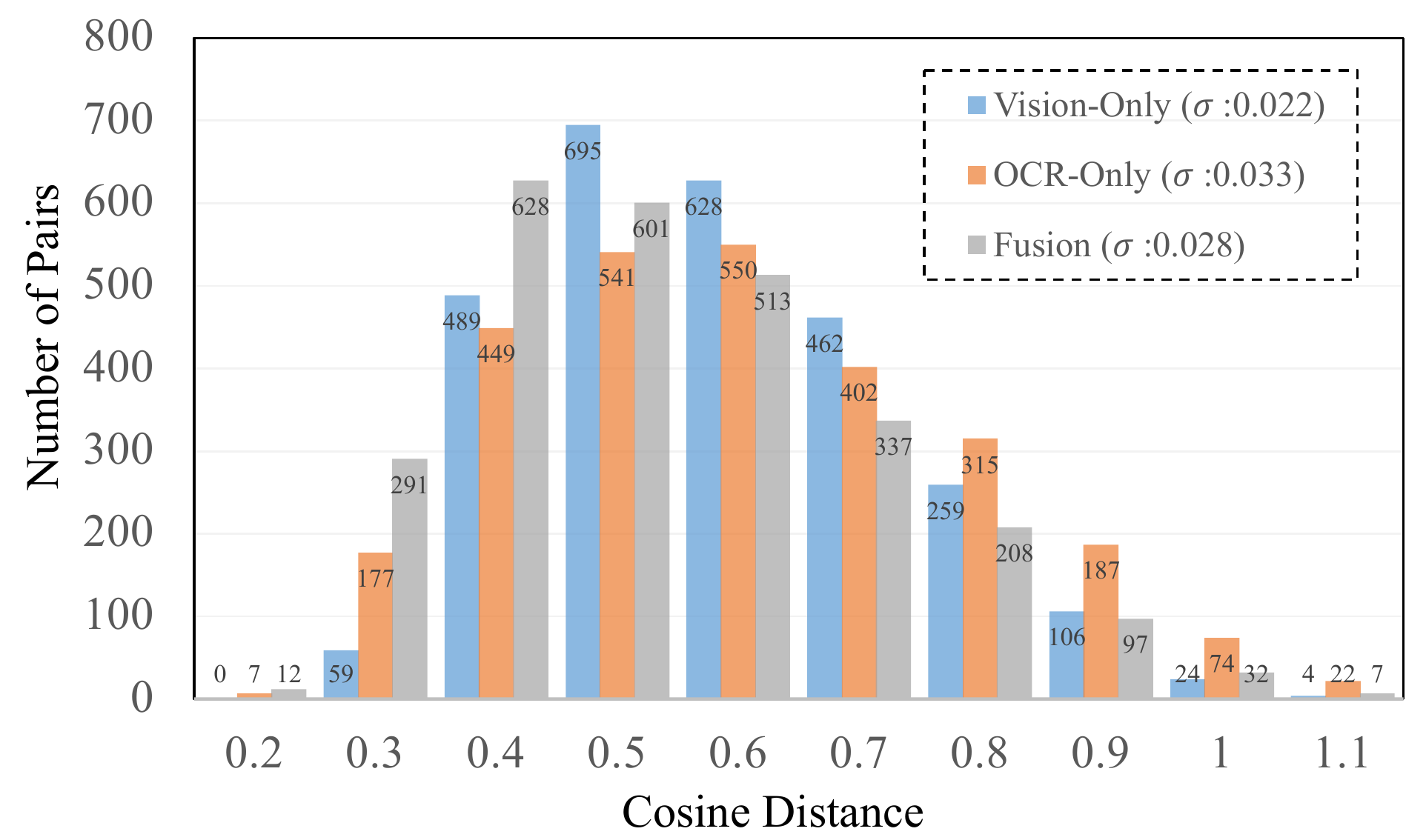}%
\label{fig:3c}	}
\caption{\textbf{Effect of Vision and Text Representation.} The matrix of cosine distance is transferred to $10 \times 10$ with linear interpolation. We only keep one decimal for Fig~\ref{fig:3c}. Cosine distance of text representation tends toward extremes, due to the more accurate semantics. (b) Similarity Matrix for Text-only. (c) Cosine Distance Distribution of Positive Pairs.}
\label{fig_sim}
\end{figure*}

\begin{table}[t]
    \centering
    \small 
    \setlength{\tabcolsep}{2mm}
    \input{table/table6}
    \caption{\textbf{Impact of Video OCR accuracy} .}
    \label{tab:table6}
\end{table}
\textbf{Ablation for \starvr.}
Table~\ref{tab:table5} shows the ablation study for \starvr.
When the text/OCR token is not used, it is set to empty set $\emptyset$ as described in Sec.~\ref{OCR token}.
With OCR tokens, \starvr achieves obvious improvements, $5.7\%$ R@1 and $10$ MdR for language-to-video retrieval, which proves the effectiveness of text semantic representation. 
And the combination of three features~(\ie{} vision, OCR semantic, and spatial representation) presents the best performance, with $16.5$ R@1 and $13.0$ MdR for language-to-video retrieval task.

\textbf{Accuracy of different Video OCR model.}
To further explore the relationship between video OCR accuracy and retrieval performance, we proposed to adopt one new metric, \ie{} recall of correlation.
Give a sentence query $S$ from ground truth and a set of video OCR tokens $\mathcal{O}_t$ = $\{T_1,T_2,...,T_n\}$ for the same video,
we firstly split the sentence $S$ to a set of words $\mathcal{O}_w$ = $\{w_1,w_2,...,w_m\}$.
If existing $w_i$ $\in$ $\mathcal{O}_t$~($w_i$ represents $i$-th word in $\mathcal{O}_w$), the sentence query is viewed as true correlation with video OCR tokens.
We calculate the final recall of correlation by the proportion of the number of true correlations and the number of whole-sentence queries.
The higher recall of correlation represents that there exists more abundant mutual information between video OCR tokens and sentence queries.
Table.~\ref{tab:table6} presents the comparison of different recalls from different video ocr models.
Kwai-API~\cite{KuaiShou_api} presents highest recall with $64.5\%$, while that of TransDETR~\cite{wu2022end} is lowest with $33.1\%$.
Obviously, video OCR tokens from TransDETR~\cite{wu2022end} have more text detection and recognition errors, and loss much text information with higher false negatives.
Finally, with OCR tokens from Kwai-API, \starvr achieves the highest performance, with $16.5$ R@1 and $13.0$ MdR.

\begin{figure}[!t]
\begin{center}

\includegraphics[width=0.48\textwidth]{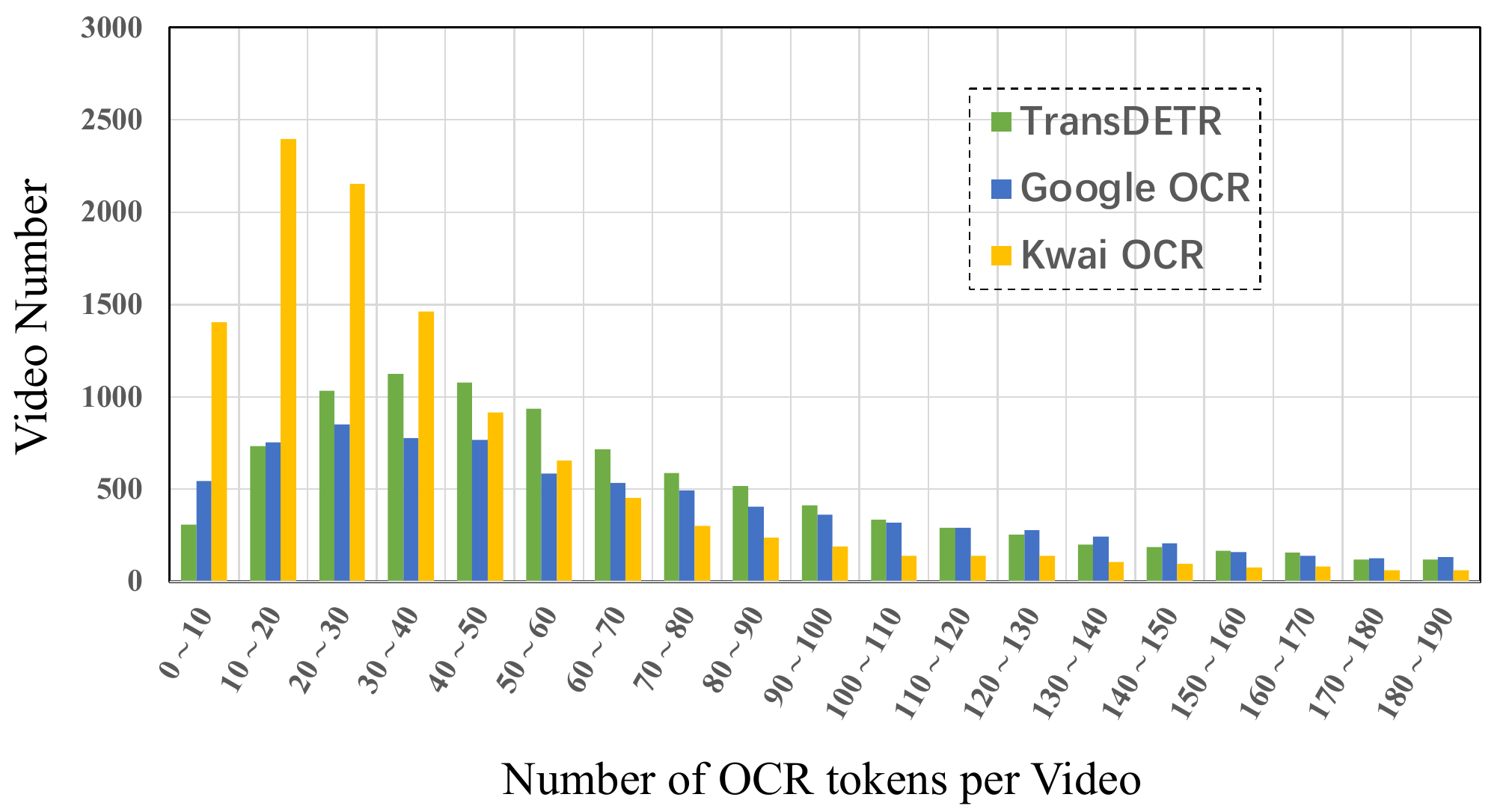}
\caption{\textbf{Distribution of OCR tokens per Video}. Accurate semantics from text/OCR tokens may cause a negative impact.}
\label{visualization_3}
\end{center}
\end{figure}

\textbf{Selection of Video OCR tokens.} 
Except for the comparison between different modal information, how to choose related text/OCR tokens also should be explored.
As shown in Fig.~\ref{visualization_3}, there is a mass of text/OCR tokens for each video, up to hundreds of text instances regardless of using which Video OCR model.
A mass of text/OCR tokens from video mainly may include two negative aspects: 1) Longer computation cost. 2) Confusing model with irrelevant information.
As shown in Fig.~\ref{visualization_2}, there are many text/OCR tokens in each video, but only a fewer video tokens are strongly related and useful for improving retrieval tasks.
A lot of OCR tokens may confuse the model, which is difficult to focus on useful information, while all OCR tokens are embedded into one final representation.
And sometimes unrelated text/OCR tokens even bring negative impact, \eg{} the description in Fig.~\ref{visualization_2}(b) contains the word `\textit{Exit}', which makes the embedding vector closer to video with the word, although the word does not represent any scene text in the video. 
To further explore the impact of text/OCR tokens, we present a simple experiment for training with only Top $K$ Video OCR tokens, which are selected by different confidence thresholds.
As shown in Table.~\ref{tab:table4}, using more Video OCR tokens can bring more performance improvements.

\begin{table*}[t]
    \centering
    \small 
    \setlength{\tabcolsep}{2mm}
    \input{table/table2}
    \caption{\textbf{Performance results on \textvr for language-and-video retrieval.} \demph{In gray} denotes initialization with CLIP weight~\cite{radford2021learning}~(\ie{} Clip4Clip) or pre-trained model on extra datasets~(\ie{} WebVid-2M).}
    \label{tab:icdar15}
\end{table*}

\textbf{Cross Domain Evaluation.} Table~\ref{tab:table5} presents the cross evaluation for different datasets.
The model trained on existing datasets~(\ie{} MSR-VTT, YouCook2) can not achieve satisfactory performance~(only $2.8$ R@1 and $134.0$ MdR for best performance) on \textvr.
Even if using the ground truth of subtitle semantics from YouCook2, the cross-domain performance still is quite low.
This is reasonable from two points: 1) Huge domain gap between existing datasets and \textvr. 2) Subtitles and scene text represent different text semantics, the model can not learn transferable representation from the subtitle.

\begin{table}[t]
    \centering
    \small 
    \setlength{\tabcolsep}{2mm}
    \input{table/table4}
    \caption{\textbf{Impact of Top $K$ Video OCR tokens.} .}
    \label{tab:table4}
\end{table}

\textbf{Text/OCR Representation $v.s$ Visual Representation.}
Table.~\ref{tab:table3} presents a interesting conclusion, \textit{text/ocr representation improves R@1 and MdR obviously, but fail for MnR.} 
To further explain the result, we provide a matrix of cosine distance for different modal representations in Fig~\ref{fig:3a} and Fig~\ref{fig:3b}, cosine distance distribution of positive pairs in Fig~\ref{fig:3c}.
Compare to visual representation, text/ocr presentation tend toward extreme, where cosine distances of positive pairs usually tend to be bigger or smaller~(smallest: $0.4$ $v.s$ $0.32$, biggest: $0.65$ $v.s$ $0.84$), as shown in Fig~\ref{fig:3a} and Fig~\ref{fig:3b}.
The whole cosine distance distribution in Fig~\ref{fig:3c} also proves the conclusion, extreme cosine distances of text/ocr representations show a higher proportion than that of visual representation~($177$ $v.s$ $59$ for $0.3$ cosine distance, $74$ $v.s$ $24$ for $1.0$ cosine distance).
Besides, we also calculate the variance for cosine distances, \ie{} $0.022$ $v.s$ $0.033$ for visual and text/OCR representations.
Due to the accurate semantics, text/OCR representation usually makes the model more confident for its prediction, which causes a more precise match and more hard confusion.
This is why Mean Rank~(Rnk) can not obtain an obvious gain, while R@1 and Median Rank~(MeR) show a significant improvement.

\subsection{Comparison with State-of-the-arts}
Table.~\ref{tab:icdar15} shows the performance of cross-modal language-and-video retrieval on \textvr.
Compare with the previous methods~(\eg{} Frozen), with fusion embedding of text semantic and visual representations, our \textvr shows obvious improvement with up to $9.1$ R@1 and $13.0$ MdR~(using Kwai-API).
On the whole, Kwai Video OCR API~\cite{KuaiShou_api} shows better performance than that of TransDETR~\cite{wu2022end} and Google VideoOCR API~\cite{Google_api}, while TransDETR~\cite{wu2022end} presents worst results.
This is reasonable that we just using the weight of TransDETR on ICDAR2015 video~\cite{karatzas2015icdar}, where only $24$ videos are used for training.
But Kwai Video OCR API~\cite{KuaiShou_api} and Google VideoOCR API~\cite{Google_api} are applicable in business, both are trained with extremely large amounts of private data.

\subsection{Conclusion}
Cross-modal video-and-language with reading comprehension is a novel challenging task requiring models to read text, relating to visual context to retrieve video.
To promote the task, we establish a large, large cross-modal video retrieval benchmark with reading comprehension, including $42.2k$ sentence queries for $10.5k$ videos of $8$ scenario domains. 
Compare with the existing benchmarks, the proposed \textvr mainly contains three advantages: sentence query containing text semantics, videos containing abundant scene text, and high resolution.

Experimental results show that current video retrieval models cannot utilize the semantics from text/OCR tokens effectively, which causes an unsatisfactory performance on our \textvr.
By contrast, our \starvr achieves a huge improvement with up to $9.1$ R@1 and $13.0$ MdR, while the fusion embedding between visual and scene text representations is used. 
We hope the \textvr will promote the development of  cross-modal video retrieval in the community.

\section{Acknowledgements} This work is supported by the National Key Research and Development Program of China (2022YFC3602601), and the Key Research and Development Program of Zhejiang Province of China (2021C02037).

\appendix


\begin{figure*}[!t]
\begin{center}

\includegraphics[width=1\textwidth]{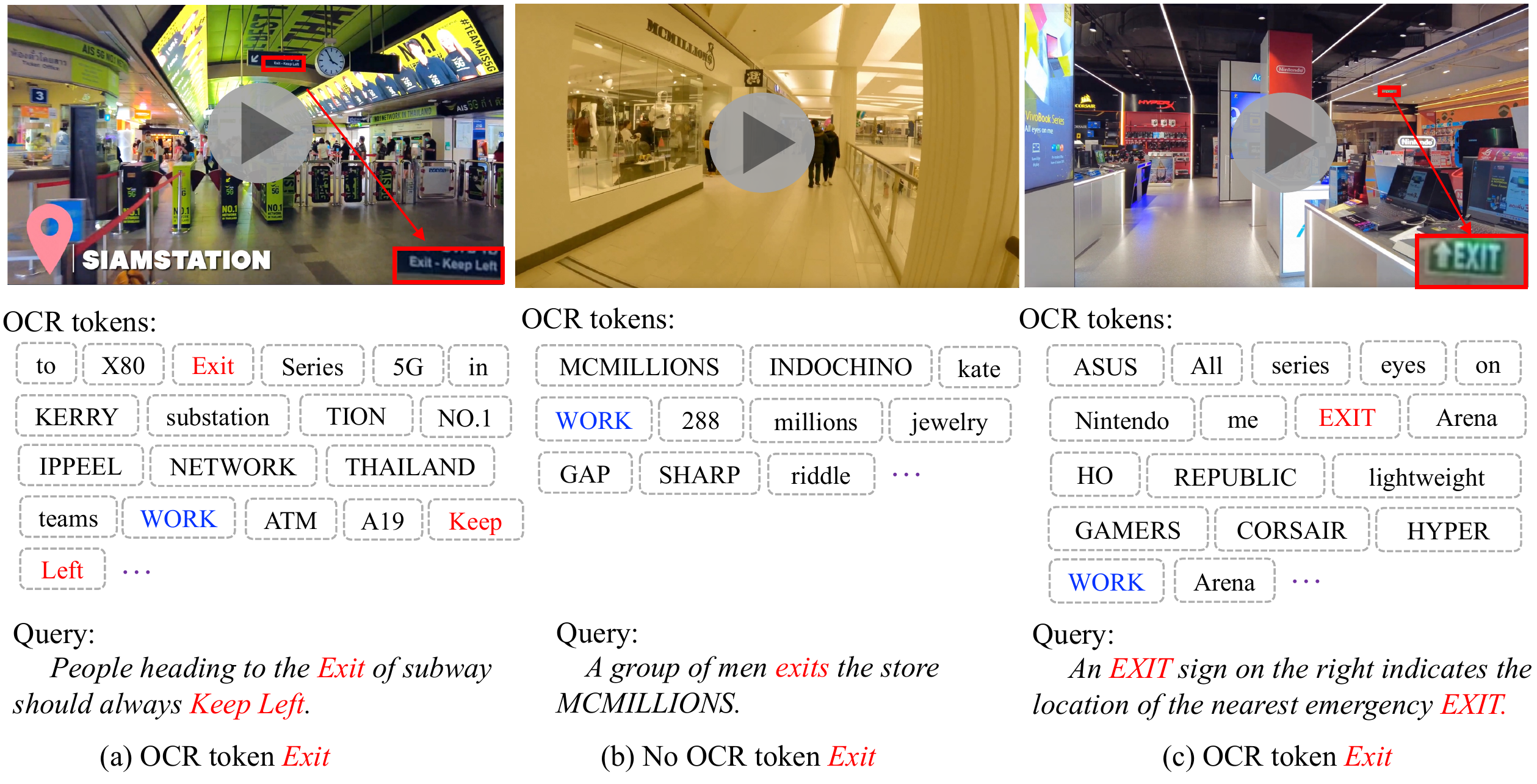}
\caption{\textbf{Same text/OCR tokens}. Accurate semantics from text/OCR tokens may cause a negative impact. \textcolor[RGB]{255,0,0}{In red} and \textcolor[RGB]{0,0,255}{in blue} refer to the related text and confused noise, respectively.}
\label{visualization_2}
\end{center}
\end{figure*}

%
%
%
%

\subsection{Illustration of most frequent words in TextVR}
\begin{figure*}[!t]
\centering
\subfloat[]{\includegraphics[width=2.0in]{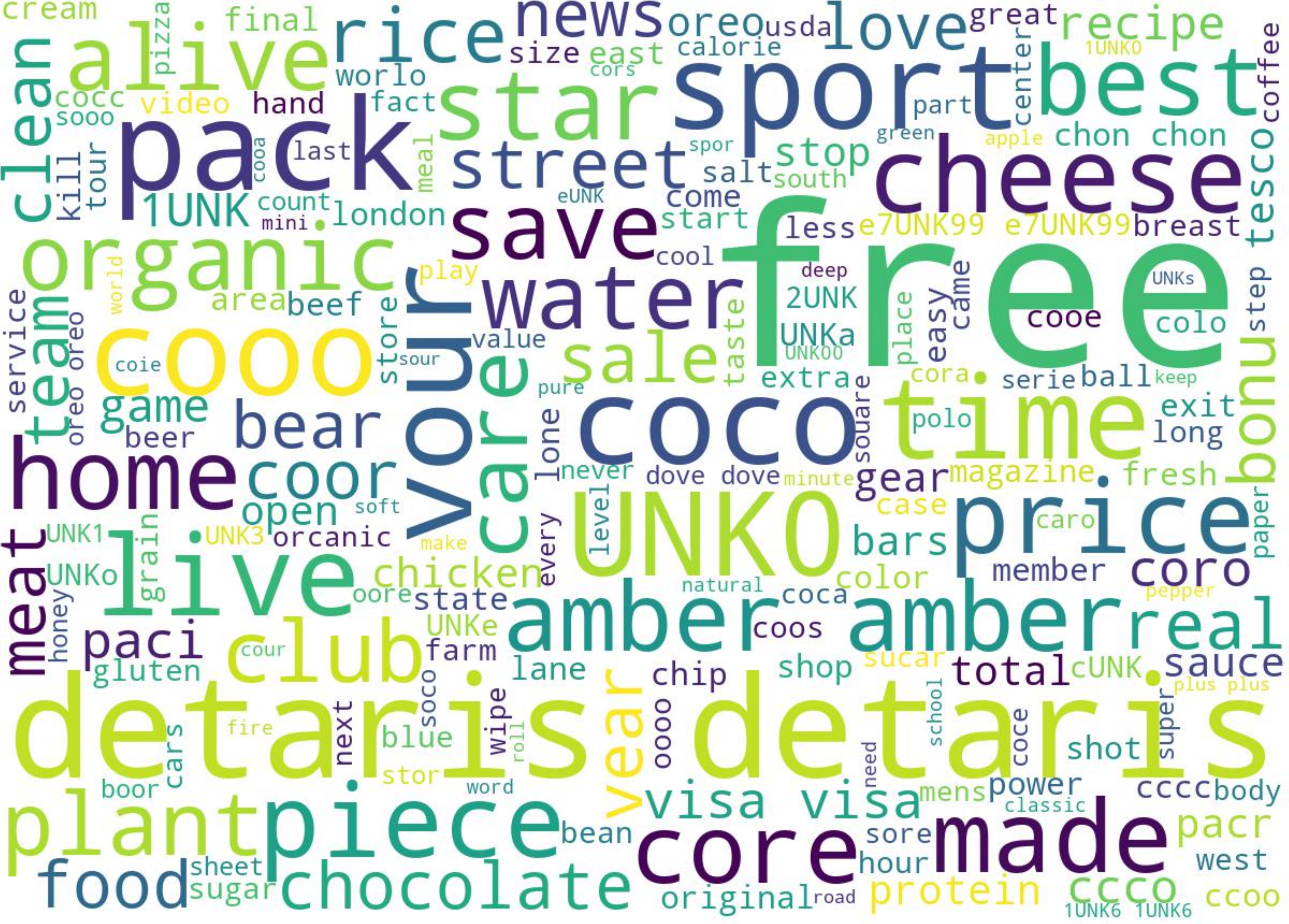}%
\label{fig:32a}}
\hfil
\subfloat[]{\includegraphics[width=2.0in]{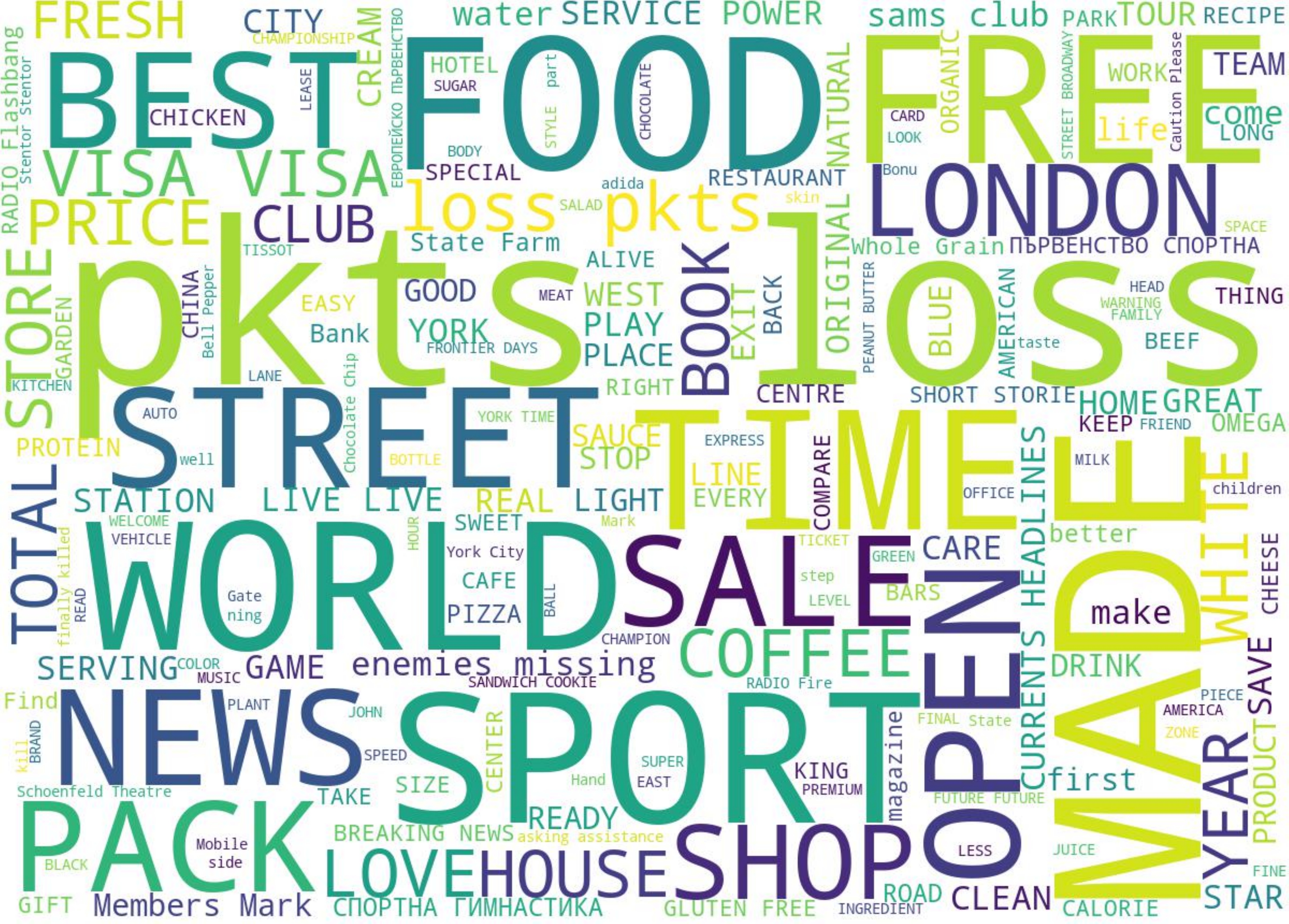}%
\label{fig:33b}}
\hfil
\subfloat[]{\includegraphics[width=2.0in]{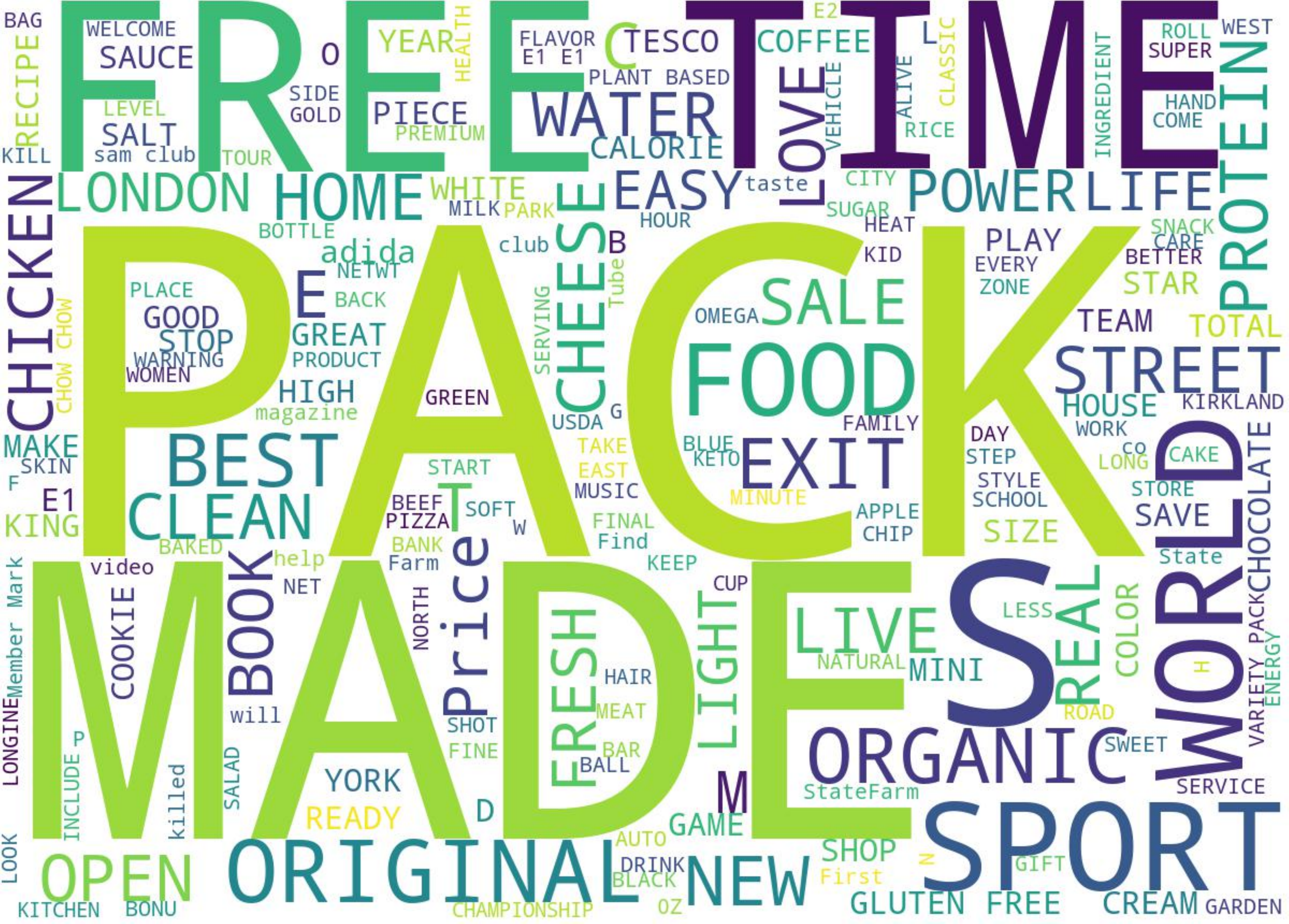}%
\label{fig:33c}	}
\hfil
\subfloat[]{\includegraphics[width=2.8in]{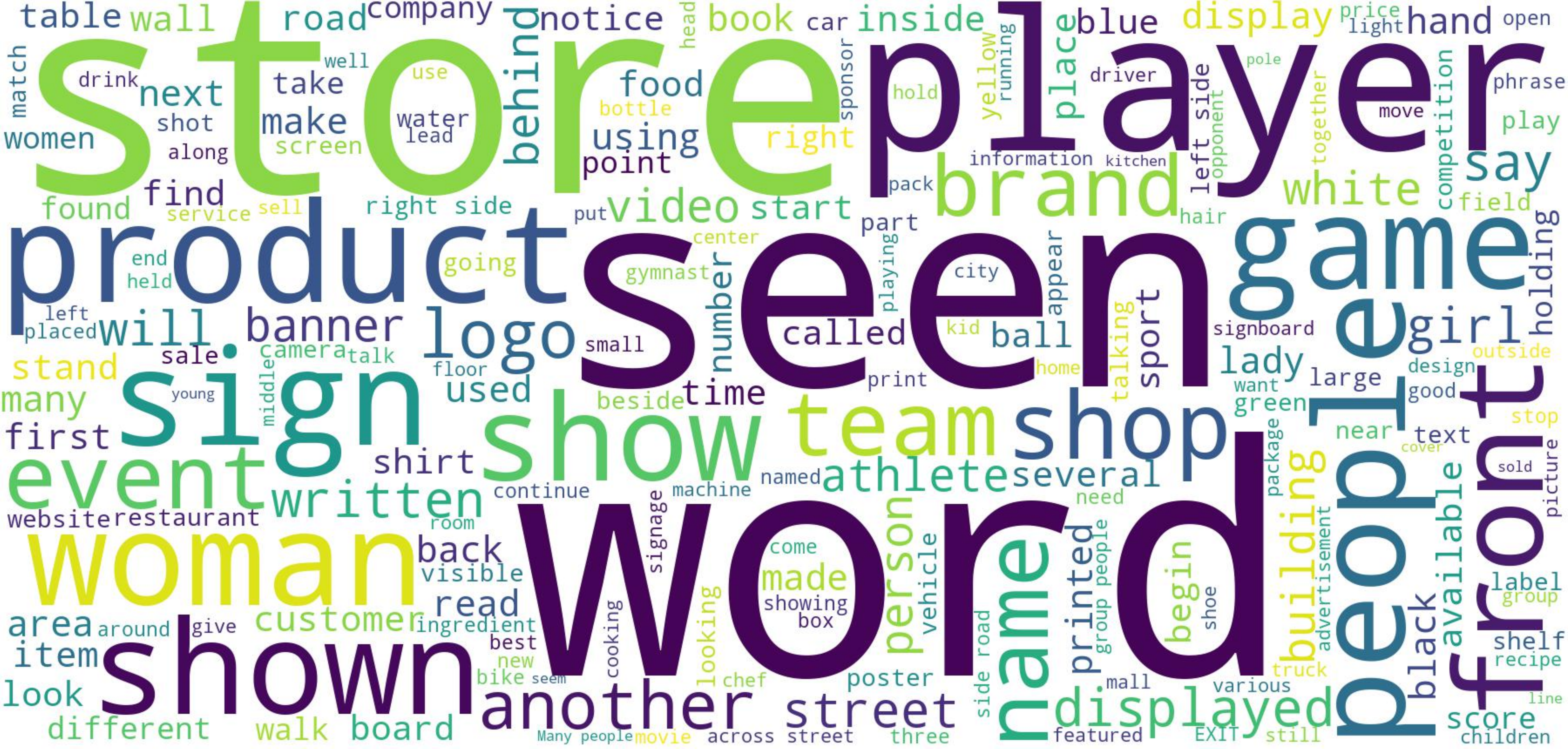}%
\label{fig:33d}	}
\hfil
\subfloat[]{\includegraphics[width=2.8in]{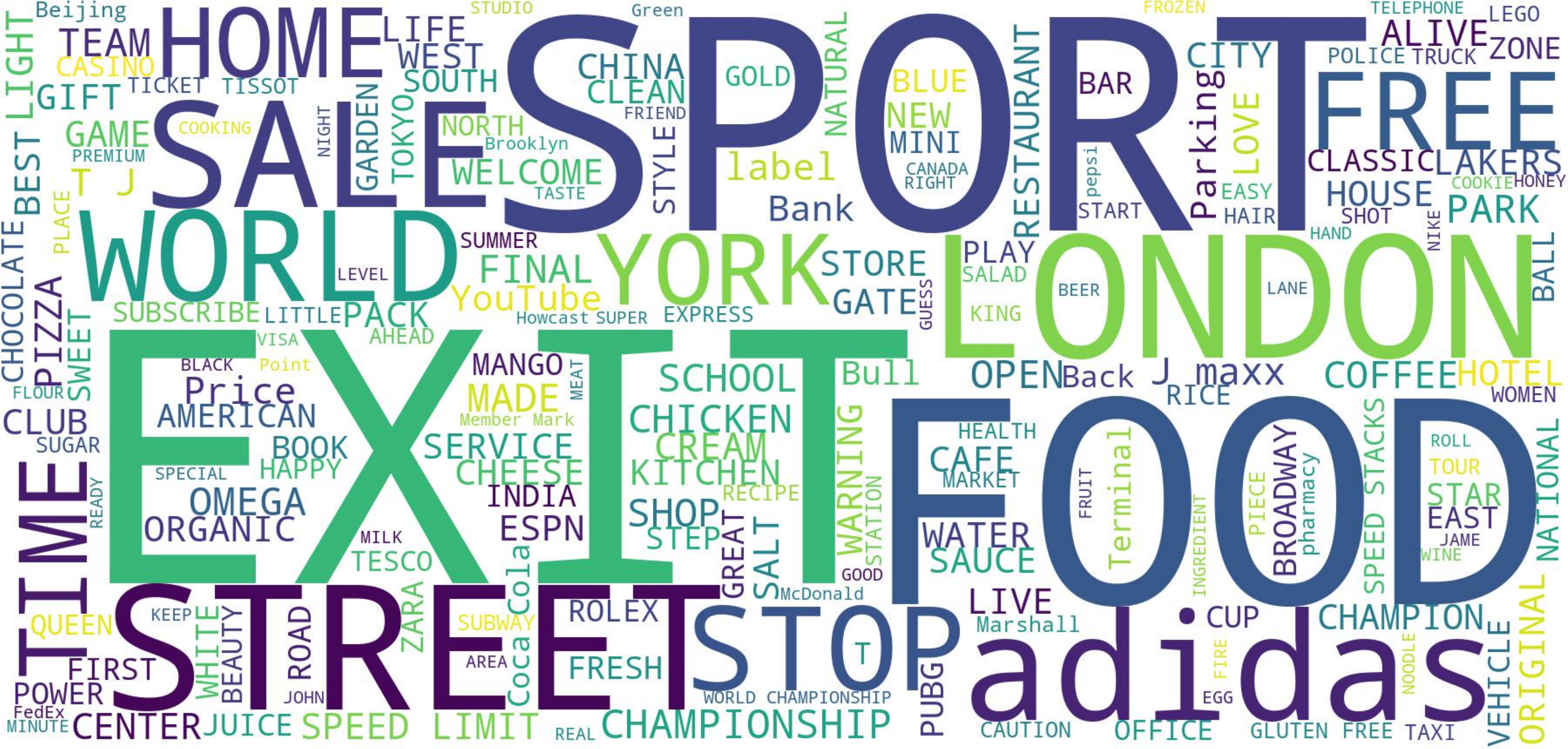}%
\label{fig:33e}	}
\caption{\textbf{Wordcloud visualizations.} The related Video OCR tokens in Fig.~\ref{fig:33e} is selected by judging whether the word~\cite{KuaiShou_api} existing in corresponding description query. (a) Wordcloud of Video OCR tokens from TransDETR~\cite{wu2022end}. (b) Wordcloud of Video OCR tokens from Google Video OCR API~\cite{Google_api}. (c) Wordcloud of Video OCR tokens from Kwai Video OCR API~\cite{KuaiShou_api} (d) Wordcloud of Sentence Queries. (e) Wordcloud of \textbf{Related} VideoOCR tokens.}
\label{ablation_semantic_visual33}
\end{figure*}

The word clouds for three Video OCR tokens, sentence queries, and related Video OCR tokens are provided in Fig.~\ref{ablation_semantic_visual33}. 
It can be seen that different Video OCR models give different frequent word results, but the main cause may be the difference of training data.
And some same words also can be found, \ie{} `SPORT', `WORLD', which also appear in the corresponding description.
The word cloud from TransDETR~\cite{wu2022end} seems more sparse, and the frequency of words is close, while that of other VideoOCR models is more discriminative.

We also visualize the word cloud for sentence queries and \textbf{related} VideoOCR tokens in Fig~\ref{fig:33d} and Fig~\ref{fig:33e}.
All words from a sentence must contain at least $4$ characters, we consider the words less than four characters usually insignificant, \eg{} `is', `you'.
As shown in Fig.~\ref{fig:33e}, it can be observed that great use of words like ‘Exit’, ‘Sport’, and ‘FOOD’, which annotators used to incorporate text tokens into the captions, which also highly frequently appear in street view scenarios.

\begin{figure}[!t]
\begin{center}

\includegraphics[width=0.48\textwidth]{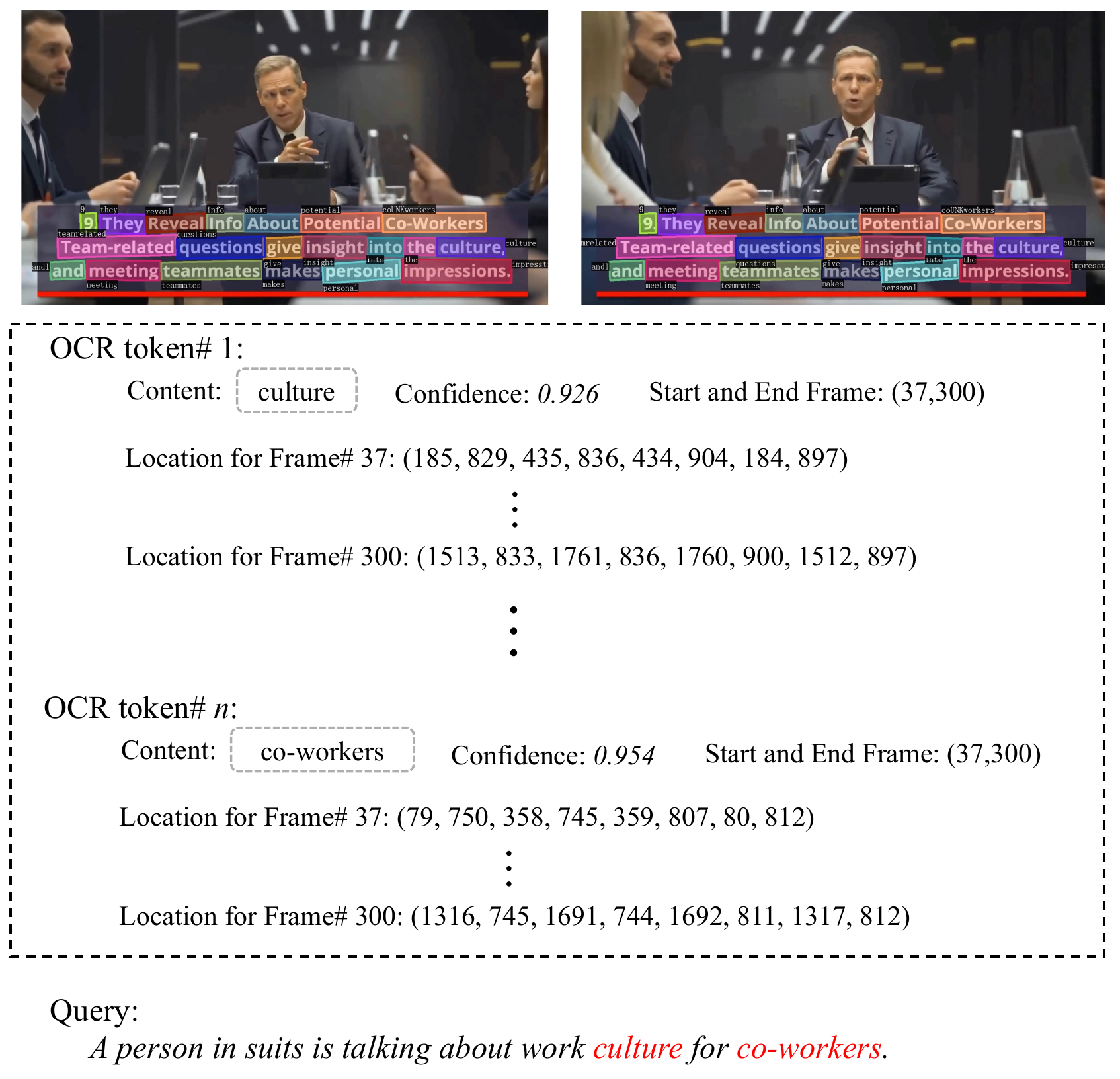}
\caption{\textbf{Visualization for Retrieval Result and Video OCR model Output}. The above text/OCR tokens information are embedded into text space-time and scene text encoders to enhance the final representation.}
\label{visualization_4}
\end{center}
\end{figure}

\textbf{Visualization for Retrieval Result.} 
To further present the form of text/OCR token used in our \starvr, we visualize one example in Fig.~\ref{visualization_4}.
We obtain text content, confidence, start and end time, and location per frame for each text instance from the Video OCR model, and embed them to enhance the final representation.

\bibliographystyle{IEEEtran}
\bibliography{main}

\end{document}

%% file: math_commands.tex
\renewcommand{\Sigma}{\mathfrak{S}}

\def\eqref#1{equation~\ref{#1}}

\def\1{\bm{1}}

\DeclareMathAlphabet{\mathsfit}{\encodingdefault}{\sfdefault}{m}{sl}
\SetMathAlphabet{\mathsfit}{bold}{\encodingdefault}{\sfdefault}{bx}{n}

\DeclareMathOperator*{\argmax}{arg\,max}
\DeclareMathOperator*{\argmin}{arg\,min}

%% file: table/table1.tex
\def\x{{$\footnotesize \times$}}
\scriptsize  
\setlength{\tabcolsep}{2.0pt}
\begin{tabular}{l|c|c|c|c|p{0.2\columnwidth}|p{0.6\columnwidth}}
    \whline
    
	Dataset & Channel & Query~(class, \# number)  & Videos & Frames & Resolution & Supported Scenario~(Video Source)   \cr\shline \hline
	\hline

	\multicolumn{7}{l}{\textit{Image Retrieval with Text Reading} }
      \\
    \hline
    SVT~\cite{wang2011end}& Text & Word, 427.0 & - & 349.0 & 720p & Street View\\
    IIIT STR~\cite{mishra2013image} & Text & Word, 50.0 & - & 10.0k & 720p &  Street View \\
    COCO-Text~\cite{veit2016coco}& Text & Word, 500.0 & - & 7.1k & 360p &  Street View \\
    COCO-Text Cap~\cite{mafla2021stacmr}& Vision \& Text & Sentence, 53.0k & - & 10.6k & - & Street View \\
    \hline
    \multicolumn{7}{l}{\textit{Video Retrieval with Vision only} }
      \\
    \hline
 
    MSR-VTT~\cite{xu2016msr} &Vision& Sentence, 200.0k &  10.0k & 4.1m  &240p & Music, Gaming, Sports,  TV shows, ...\\
    
    MSVD~\cite{chen2011collecting} &Vision& Sentence, 120.0k &  2.0k & 542.0k  & 360p & YouTube\\
    
    VATEX-EN-R~\cite{wang2019vatex}&Vision & Sentence, 349.9k &  35.0k & 6.1m &360p & Human Activity\\
    
    ActivityNet~\cite{caba2015activitynet}&Vision & Sentence, 100.0k &  100.0k & 150.0m& 480p & Human Activity\\
    \hline
    \multicolumn{6}{l}{\textit{Video Retrieval with Vision and Text aggregation} }
      \\
    \hline
    YouCook2~\cite{zhou2018towards}&Vision \& Subtitle & Sentence, 13.8k &  13.8k & 6.8m  & 720p & Cooking
    \cr\cline{7-7} 
    \textbf{\textvr}~(ours) & \textbf{Vision \& Text} & \textbf{Sentence}, 42.2k & 10.5k & 6.4m & \textbf{1080p ($72\%$)}, 720p ($19\%$), others ($9\%$) & $8$ Categories: Street View~(Indoor), Street View~(Outdoor), Game, Sports,  Driving, Activity, TV Show, Cooking.\\
    
     \whline
\end{tabular}

%% file: table/table3.tex
\def\x{{$\footnotesize \times$}}
\scriptsize
\setlength{\tabcolsep}{3pt}
\begin{tabular}{c|c|ccccc|ccccc}
    \whline
	
	\multirow{2}*{\makecell[c]{Dataset}} &	\multirow{2}*{\makecell[c]{Input Channel}} &
	\multicolumn{5}{c|}{Language-to-Video}&
	
	\multicolumn{5}{c}{Video-to-Language}   \cr\cline{3-12}  
	
	
    ~ & ~ & R@1$\uparrow$  & R@5$\uparrow$ & R@10$\uparrow$ & MdR$\downarrow$ & MnR$\downarrow$ & R@1$\uparrow$  & R@5$\uparrow$ & R@10$\uparrow$ & MdR$\downarrow$ & MnR$\downarrow$ \cr\shline \hline
	\hline
    \multirow{3}*{\makecell[c]{MSR-VTT}~\cite{xu2016msr}} & Vision-only & 19.5 & 45.2 & 58.0 & 6.0 & 37.2 & 21.1 & 48.0 & 58.5 & 6.0 & 32.8 \\
     & Text-only  & 3.4 & 10.1 & 16.0 & 30.1 & 213.8 & 2.1 & 7.6 & 13.2 & 188.5 & 281.3 \\
     & Vision \& Text & 20.0~(\green{+0.5}) & 46.9 & 58.8 & 5.7~(\green{-0.3}) & 37.0 & 21.2 & 48.1 & 59.6 & 6.0 & 32.1 \\
    \hline
    
    \multirow{5}*{\makecell[c]{YouCook2}~\cite{zhou2018towards}} & Vision-only & 12.8 & 32.8 & 44.1 & 14.0 & 75.0 & 10.9 & 31.6 & 43.5 & 14.5 & 76.9 \\
    & Subtitle-only & 25.2 & 47.2 & 55.2 & 7.0 & 93.1 & 23.2 & 45.4 & 54.5 & 7.0 & 104.4 \\
    & Text-only & 1.9 & 5.8 & 7.9 & 385.0 & 484.7 & 1.8 & 4.6 & 6.5 & 377.0 & 468.0 \\
    & Vision \& Subtitle & 32.5 & 56.1 & 65.7 & 4.0 & 32.1 & 30.2 & 57.1 & 67.7 & 4.0 & 32.9 \\
     & Vision \& Text & 13.1~(\green{+0.3}) & 32.6 & 44.1 & 16.0~(\green{+2.0}) & 79.5 & 12.1 & 30.3 & 42.7 & 16.5 & 82.1 \\
    \hline
    
    \multirow{3}*{\makecell[c]{\textvr}} & Vision-only & 7.4 & 22.7 & 32.5  & 26.0 & 132.3 & 7.4 & 22.3 & 33.7 & 25.0 & 125.4 \\
    & Text-only & 11.9 & 29.0 & 38.0 & 24.0 & 215.9 & 12.2 & 30.2 & 39.8 & 22.0 & 208.0 \\
     & Vision \& Text &
     16.5~(\green{+9.1}) & 37.3 & 47.3 & 13.0~(\green{-13.0}) & 124.2 & 16.0 & 38.0 & 49.0 & 11.0 & 112.8\\
     \whline 
\end{tabular} 

%% file: table/table5.tex
\def\x{{$\footnotesize \times$}}
\scriptsize
\setlength{\tabcolsep}{3pt}
\begin{tabular}{l|c|ccccc|ccccc}
    \whline
	
	\multirow{2}*{\makecell[c]{Method}} &
	\multirow{2}*{\makecell[c]{Trained on}} &
	\multicolumn{5}{c|}{Language-to-Video}&
	
	\multicolumn{5}{c}{Video-to-Language}   \cr\cline{3-12}  
    ~ & ~ &  R@1$\uparrow$  & R@5$\uparrow$ & R@10$\uparrow$ & MdR$\downarrow$ & MnR$\downarrow$ & R@1$\uparrow$  & R@5$\uparrow$ & R@10$\uparrow$ & MdR$\downarrow$ & MnR$\downarrow$ \cr\shline \hline
	\hline
    \starvr~(w/o OCR tokens) & \textvr & 7.4 & 22.7 & 32.5 & 26.0 & 132.3 & 7.4 & 22.3 & 33.7 & 25.0 & 125.4 \\
    \starvr~(OCR semantic) & \textvr & 13.1 & 33.8 & 42.7 & 16.0 & 127.9 & 13.5 & 34.7 & 44.6 & 16.0 & 115.2 \\
    \starvr~(OCR semantic\&spatial) & \textvr & 16.5 & 37.3 & 47.3 & 13.0 & 124.2 & 16.0 & 38.0 & 49.0 & 11.0 & 112.8  \\
    \hline
    \starvr~(OCR semantic\&spatial) & MSR-VTT~\cite{xu2016msr} & 2.8 & 8.4 & 13.4 & 134.0 & 352.9 & 2.4 & 8.3 & 12.9 & 14.1 & 337.3 \\
    \starvr~(OCR semantic\&spatial) &  YouCook2~\cite{zhou2018towards} & 0.8 & 1.9 & 2.8 & 1011.0 & 1023.1 & 0.6 & 1.6 & 2.2 & 1102.0 & 1201.0  \\
    \starvr~(Subtitle semantics) &  YouCook2~\cite{zhou2018towards} & 1.2 & 2.8 & 4.3 & 906.0 & 1011.2 & 1.8 & 3.1 & 4.9 & 967.0 & 1023.1  \\
    \starvr~(OCR semantic\&spatial) & \textvr & 16.5 & 37.3 & 47.3 & 13.0 & 124.2 & 16.0 & 38.0 & 49.0 & 11.0 & 112.8 \\
    
     \whline 
\end{tabular} 

%% file: table/table6.tex
\def\x{{$\footnotesize \times$}}
\scriptsize
\setlength{\tabcolsep}{3pt}
\begin{tabular}{l|c|ccccc}
    \whline
	
	\multirow{2}*{\makecell[c]{Video OCR}} &
	\multirow{2}*{\makecell[c]{ Recall/\%}} &
	\multicolumn{5}{c}{Language-to-Video} \cr\cline{3-7}  
    ~ & ~ &  R@1$\uparrow$  & R@5$\uparrow$ & R@10$\uparrow$ & MdR$\downarrow$ & MnR$\downarrow$  \cr\shline \hline
	\hline
    TransDETR~\cite{wu2022end}& 33.1 & 9.1 & 23.8 & 34.2 & 24.0 & 150.1 \\
    Google-API~\cite{KuaiShou_api} & 63.4 & 15.4 & 34.9  & 44.2 & 16.0 & 111.8  \\
    Kwai-API~\cite{KuaiShou_api} & 64.5 & 16.5 & 37.3  & 47.5 & 13.0 & 124.2  \\

     \whline 
\end{tabular} 

%% file: table/table2.tex
\def\x{{$\footnotesize \times$}}
\scriptsize
\setlength{\tabcolsep}{3pt}
\begin{tabular}{c|c|c|ccccc|ccccc}
    \whline
	
	\multirow{2}*{\makecell[c]{Method}} &	\multirow{2}*{\makecell[c]{Video OCR}} &\multirow{2}*{\makecell[c]{Pre-train}} &
	\multicolumn{5}{c|}{Language-to-Video}&
	
	\multicolumn{5}{c}{Video-to-Language}   \cr\cline{4-13}  
	
	
    ~ & ~ &~ & R@1$\uparrow$  & R@5$\uparrow$ & R@10$\uparrow$ & MdR$\downarrow$ & MnR$\downarrow$ & R@1$\uparrow$  & R@5$\uparrow$ & R@10$\uparrow$ & MdR$\downarrow$ & MnR$\downarrow$ \cr\shline \hline
	\hline
	 \multicolumn{13}{l}{\textit{Input Channel: Vision-Only} }
    \\
    \hline
    ClipBERT~\cite{lei2021less} & - &- &  6.3  & 16.3 & 26.7 & 30.0 & 183.3 & 5.3 & 18.2 & 28.6 & 29.0 & 170.4\\
    Frozen~\cite{bain2021frozen} & - &- &  7.4  & 22.7 & 32.5 & 26.0 & 132.3 & 7.4 & 22.3 & 33.7 & 25.0 & 125.4\\
	 our \starvr & - &- & 7.4  & 22.7 & 32.5 & 26.0 & 132.3 & 7.4 & 22.3 & 33.7 & 25.0 & 125.4\\
	 \hline
	 \multicolumn{13}{l}{\textit{Input Channel: Text-Only} }
    \\
    \hline
	 our \starvr & TransDETR~\cite{wu2022end} & - & 3.9 & 11.7  & 16.0 & 210.0 & 451.8 & 2.6 & 10.4  & 15.8 & 179.0 & 486.2 \\
	 our \starvr & Google-API~\cite{KuaiShou_api} &- & 10.1 & 22.2 & 28.0 & 84.0 & 342.8 & 8.8 & 20.1 & 26.7 & 81.0 & 360.6 \\
	 our \starvr & Kwai-API~\cite{KuaiShou_api} & - &11.9 & 29.0 & 38.0 & 24.0 & 215.9 & 12.2 & 30.2 & 39.8 & 22.0 & 208.0 \\
	  \hline
	 \multicolumn{12}{l}{\textit{Input Channel: Vision \& Text} }
    \\
    \hline
	 our \starvr & TransDETR~\cite{wu2022end} &- & 9.1 & 23.8 & 34.2 & 24.0 & 150.1 & 9.2 & 26.0 & 34.5 & 25.0 & 131.4\\
	 our \starvr & Google-API~\cite{KuaiShou_api} &- & 15.4 & 34.9  & 44.2 & 16.0 & \textbf{111.8} & 15.2 & 35.1 & 44.3 & 15.4 & \textbf{106.2} \\
	 our \starvr & Kwai-API~\cite{KuaiShou_api} &- & \textbf{16.5} & \textbf{37.3}  & \textbf{47.3} & \textbf{13.0} & 124.2 & \textbf{16.0} & \textbf{38.0}  & \textbf{49.0} & \textbf{11.0} & 112.8
	 \\
	 \demph{our \starvr} & \demph{Kwai-API~\cite{KuaiShou_api}} & \demph{WebVid-2M~\cite{bain2021frozen}} & \demph{19.2} & \demph{38.2}  & \demph{47.5} & \demph{13.0} & \demph{138.2} & \demph{18.6} & \demph{40.1}  & \demph{48.8} & \demph{12.0} & \demph{132.8}
	 \cr\hline 

     \whline 
\end{tabular} 

%% file: table/table4.tex
\def\x{{$\footnotesize \times$}}
\scriptsize
\setlength{\tabcolsep}{3pt}
\begin{tabular}{c|ccccc}
    \whline
	
	\multirow{2}*{\makecell[c]{Text/OCR Tokens}} &
	\multicolumn{5}{c}{Language-to-Video}  \cr\cline{2-6}  
    ~ &  R@1$\uparrow$  & R@5$\uparrow$ & R@10$\uparrow$ & MdR$\downarrow$ & MnR$\downarrow$  \cr\shline \hline
	\hline
    Top 10 &  13.6 & 33.7 & 42.1 & 17.0 & 143.6  \\
    Top 20 &  14.9 & 36.7 & 45.8 & 14.0 & 126.7 \\
    All & 16.5 & 37.3 & 47.3 & 13.0 & 124.2  \\
     \whline 
\end{tabular} 